%% file: main.tex
\documentclass{article}

\PassOptionsToPackage{numbers,compress}{natbib}
 \usepackage[preprint]{neurips_2026}

\usepackage[utf8]{inputenc} 
\usepackage[T1]{fontenc}    
\usepackage{hyperref}       
\usepackage{url}            
\usepackage{booktabs}       
\usepackage{amsfonts}       
\usepackage{nicefrac}       
\usepackage{microtype}      
\usepackage{xcolor}

\definecolor{darkblue}{rgb}{0, 0, 0.5}
\hypersetup{colorlinks=true, citecolor=darkblue, linkcolor=darkblue, urlcolor=darkblue}

\usepackage{caption}
\usepackage{quoting}
\usepackage{inconsolata}
\usepackage{graphicx}
\usepackage{xspace}
\usepackage[subtle]{savetrees}
\usepackage[most]{tcolorbox}
\usepackage{colortbl}
\usepackage{microtype}
\usepackage{xcolor}
\usepackage{multirow}
\usepackage{multicol}
\usepackage{longtable}
\usepackage{listings}
\usepackage{enumitem}
\usepackage{psfrag}
\usepackage{verbatim}
\usepackage{pifont}
\usepackage{cleveref}
\usepackage{adjustbox}
\usepackage{makecell}
\usepackage{subcaption}
\usepackage{wrapfig}
\usepackage{nicefrac}
\usepackage{mathtools}
\usepackage{amsfonts}
\usepackage{amsmath}
\usepackage{amssymb}
\usepackage{amsthm}
\usepackage{mathrsfs}
\usepackage{ragged2e}
\usepackage{courier}
\usepackage{tabularx}

\definecolor{berkeleyblue}{HTML}{003262}
\definecolor{berkeleylightblue}{HTML}{3B7EA1}
\definecolor{berkeleygold}{HTML}{FDB515}
\definecolor{main}{HTML}{4472C4}
\definecolor{sub}{HTML}{EBF4FF}
\definecolor{sub}{HTML}{EBF4FF}

\DeclareMathSizes{10}{9}{6}{5}

\title{Pretrained Persona Mixture Models and Tandem Models for Human Simulation}

\author{Minwoo Kang$^{1}$\thanks{Correspondence:\texttt{\{minwoo\_kang\}@berkeley.edu}}, T\'ea Wright$^{1}$, Seun Eisape$^{1}$, Ayush Raj$^{1}$, Suhong Moon$^{1}$,\\
\textbf{ Joseph Suh$^{1}$, Alane Suhr$^{1\dagger}$, David M. Chan$^{1\dagger}$, John Canny$^{1 2}$\thanks{Equal advising contributions; listed in alaphabetical order.}}\\
$^{1}$University of California, Berkeley $^{2}$ Google\\
}

\begin{document}

\maketitle

\input{sections/s0_abstract}
\input{sections/s1_introduction}

\input{sections/s2_related}

\input{sections/s3_thesis}
\input{sections/s4_realism}
\input{sections/s5_dialogue_acts}

\input{sections/s6_diversity}
\input{sections/s7_challenges}
\input{sections/s8_conclusion}

\begin{ack}
Authors, as part of their affiliation with UC Berkeley, were supported in part by the National Science Foundation, US Department of War, and/or the Berkeley Artificial Intelligence Research (BAIR) industrial alliance program. 
This research was also developed with funding from the Defense Advanced Research Projects Agency (DARPA) under Contract No. FA8650-23-C-7316, an Amazon Research Award, a gift from Google, and a Technical AI Safety Research award from Coefficient Giving.
The views, opinions and/or findings expressed are those of the authors and should not be interpreted as representing the official views or policies of any sponsor, the Department of War, or the U.S. Government.
\end{ack}

\bibliographystyle{unsrtnat}
\bibliography{higgins}
\appendix
\input{sections/a2_experimental_details}
\input{sections/a3_prompts}
\input{sections/a4_generation_examples}
\input{sections/a4_metrics}
\input{sections/a5_full_results}

\end{document}

%% file: sections/s0_abstract.tex
\begin{abstract}
We argue here that the current dominant practice in LLM human simulation---prompting instruction-tuned  ``assistant'' language  models  to role-play personas---is inaccurate and produces stereotyped predictions (lacking natural diversity).
It has previously been shown that LLMs can be ``bound'' to personas using naturalistic, free-text dialog avoiding stereotyping. Here we show
that binding can also be achieved using short, individual samples of dialog from specific people. Demographics can be added later without negative effects by simply querying the model.
We use the term Persona Mixture Models (PMMs) for well-calibrated human models, currently realized as pretrained base models. We show that they produce more accurate 
predictions than instruction-tuned models and
retain more of the lexical, semantic, and pragmatic diversity found in human dialog.
We measure realism and diversity of LLMs simulating human interlocutors across a diverse set of corpora spanning open-domain text, human-AI chat, and task-oriented dialogue between human speakers.
However, base pretrained models can produce out-of-domain
dialog and may lose some of the human's internal
state over long contexts. We
propose and explore {\em tandem models} which combine a pre-trained model with an instruction-tuned supervisor. Tandem models achieve the best 
overall accuracy and diversity in our experiments. 
\end{abstract}

%% file: sections/s1_introduction.tex
\section{Introduction}
\label{sec:introduction}

Large language models (LLMs) are increasingly used as proxies for human users.
For example, recent work 
uses LLMs as \emph{simulated users} for training and evaluating LLMs: rather than recruiting human participants for every iteration, a simulated user ``interacts'' with a model in multi-turn dialogue, providing automatic feedback at scale.
Other work
in computational social science uses LLMs as synthetic survey respondents and study participants~\citep{argyle2023out, aher2023using, bail2024generative, ziems2023large}.
The practice of building classical task-oriented dialogue systems through user simulations~\cite{levin2000stochastic, georgila2006user, schatzmann2007agenda}
has continued to prove effective even as the underlying technology has moved to LLMs, for problems including cooperative planning~\citep{lin2024decision}, cooperative task completion~\citep{li2023camel, lewis2017deal}, end-to-end evaluation of dialogue agents~\citep{kazi2024llm, sekulic2024reliable, seshadri2026lost, yao2024tau, zhou2026sim2real}, and in reinforcement learning pipelines where simulator fidelity determines the quality of the reward signal~\citep{wu2026humanlm, abdulhai2025consistently, gandhi2026simulating}.

Most human simulation work using
 LLMs uses instruction-tuned (IT) ``chat'' models,
prompted with persona descriptions~\citep{ni2026survey, shanahan2023roleplay, wang2024rolellm, park2023generative, terragni2023incontext, park2024generativeagentsimulations1000}.
While such models have an easy-to-use interface (i.e., via natural language instructions) and produce plausible sounding responses, we show that prompting ALMs (Agent Language Models realized as IT models) to simulate human users has inherent problems.
Instead we argue for the use of ``PMMs'' - Persona Mixture models, currently realized as pretrained models. 
PMMs by construction are mixtures over user personas, and
those personas can be ``bound'' by appropriate context. Binding can be done either by naturalistic life stories (prior work) or by samples of dialog from a specific user (this work). 

In contrast, instruction-tuning uses post-training datasets which are highly specialized for specific tasks like multi-step planning, logical reasoning, and coding, and are designed to create a single ``voice'' for an agent. Instruction-tuning also forces the model to 
follow explicit or implicit instructions coded in each dialog turn, so that IT models no longer have a simple characterization as mixtures over human dialog contexts. Virtually all post-training moves IT models {\em away} from idealized PMMs.

The most direct effect of this is to shift the model output distribution away from the distribution of human dialog. The papers ~\citep{santurkar2023whose,moon2024virtual,kang2025deep} showed a substantial calibration loss for IT models compared to pretrained models for human simulation. Other work showed that instruction-tuning decreases generative diversity: reward maximization concentrates probability mass on stylistically uniform, assistant-coded completions~\citep{kirk2024understanding, padmakumar2024does, yun2025price, go2023aligning, jiang2025hivemind,west2025base}.
For user simulation, we want LLMs to reflect the \emph{broad} distribution of natural human utterances, including ambiguous, resistant, and off-task turns, which IT models do not. 

Critically, \emph{pretrained} base models approximate the empirical distribution of human authorship and lexical, stylistic, and pragmatic variations present in web-scale data~\citep{radford2019language, brown2020language, andreas2022language, naous2026flipping}.
Yet the inaccuracies of IT human simulators have been reported as failures of language models in general~\cite{zhou2026sim2real, cheng2026sycophantic, li2025llm, lyman2025balancing, ni2026survey, maitra2025dialogue, ivey2024real, crockett2025ai, kapania2025simulacrum, simmons2022moral} without
determining whether the difficulties were intrinsic
to LLMs, or consequences of post-training.
For example~\citep{anthis2025social} describes five
key challenges for LLM human simulation as diversity, bias, sycophancy, alienness and generalization. The first four of these are in fact
direct and expected consequences of post-training, and are not shared by pre-trained models. We discuss some updated challenges for LLM simulation later in the paper. 
We hope by way of this paper to move the field forward past these challenges to a frontier of
LLM simulation that is much closer to human
behavior.

Specifically, the claims of our paper are:
\begin{enumerate}[label=\texttt{(C\arabic*)}, itemsep=5pt]
\item\textbf{Pretrained LMs provide high-quality \emph{prediction} of user utterances from small conversation samples, and are significantly more accurate than IT models.}
If the task is to generate a plausible next user turn, the relevant target is the empirical distribution of what humans write in context.  (sec. 4).
\item\textbf{Pretrained LMs better reflect the \emph{pragmatic behaviors} of human interlocutors.} Real users are not merely stylistically diverse; they also differ yet converge at the corpus-level in what communicative acts they perform and when they perform them. (sec. 5)
 \item\textbf{Pretrained LMs better capture the \emph{diversity }of language produced by human speakers.} User simulation requires coverage; a simulator that repeatedly produces similar utterances in ``agent voice'' may appear reasonable under isolated inspection, but it is a poor model of a population of users. (sec. 6)
\end{enumerate}
The rest of the paper develops and tests these claims empirically. Section~\ref{sec:thesis} formalizes the distinction between pretrained LMs as \emph{Persona Mixture Models (PMMs)} and instruction-tuned models as assistant models. Sections \ref{sec:realism}, \ref{sec:dialogueacts}, and \ref{sec:diversity} then evaluate the three requirements of faithful user simulation: distributional fit to human utterances, reflecting the pragmatic dialogue structure exhibited by human interlocutors, and lexical and semantic diversity. Across these analyses, we substantiate that LLMs optimized to be good \emph{assistants} are not the models best suited to simulate \emph{human users}.

%% file: sections/s2_related.tex
\section{Background \& Related Work}
\label{sec:related}

\subsection{LLM-based User Simulation}
\label{sec:related-user-sim}
User simulation has a long history in dialogue research, ranging from
statistical and agenda-based simulators for spoken dialogue
systems~\cite{levin2000stochastic, georgila2006user, schatzmann2006survey, schatzmann2007agenda}
to data-driven neural approaches~\cite{kreyssig2018neural, lin2022gentus}.
In the LLM era, the dominant paradigm has shifted toward prompting
instruction-tuned (IT) models with persona descriptions or role-play
instructions~\cite{ni2026survey, shanahan2023roleplay, wang2024rolellm, terragni2023incontext, li2016persona, zhang2018personalizing, mazare2018training, tseng2024two}.
A growing body of evidence, however, suggests that this approach
inherits pathologies of assistant-aligned models that are fundamentally
at odds with the demands of user simulation: bias ~\citep{santurkar2023whose,moon2024virtual,kang2025deep},
reduced lexical and semantic diversity~\cite{kirk2024understanding, padmakumar2024does, yun2025price, jiang2025hivemind, lu2026assistant, binz2026posttrainingmakeslargelanguage},
sycophancy and over-personalization~\cite{cheng2026sycophantic, perez2023discovering, sharma2024towards},
and persona drift over multi-turn interactions~\cite{li2025llm, li2024measuring}.
Recent work has begun to surface this gap: \citet{lyman2025balancing, binz2026posttrainingmakeslargelanguage}
show that base models retain greater behavioral fidelity for
social-science applications \citet{moon2026identity}. \citet{naous2026flipping} argue that
``better assistants yield worse simulators'' but rather than exploring base models, they consider models fine-tuned for the simulation task.
Yet much current effort still goes into engineering
fixes~\citep{wu2026humanlm, abdulhai2025consistently, zhu2026dial, li2024measuring, zhang2026userlmr1modelinghumanreasoning}
for problems that could be sidestepped by a different choice of model
type~\citep{moon2024virtual, zhu2025bare, kang2025deep}.


\subsection{Evaluating User Simulators}

The evaluation of user simulators attempts to quantify the discrepancy between simulated and real-world interactions \citep{zhou2026sim2real}.
\citet{naous2026flipping} assess fine-tuned LLM user simulators
using metrics such as intent coverage via n-gram overlap, first-turn diversity, and likelihood scores from an AI-detector Pangram \citep{emi2024technicalreportpangramaigenerated}
to estimate naturalness.
\citet{wu2026humanlm} rely on an automated judge to score alignment with ground-truth human responses on their comprehensive benchmark, \textsc{humanual}.
However, recent work demonstrates that optimizing for judge-based rewards leads to reward hacking, actually decreasing the log-likelihood of ground-truth human responses \citep{gandhi2026simulating}.
Instead of automated judges,
some recent work favors statistical measures of human-likeness~\cite{zhou2026sim2real,seshadri2026lost}.,
A recurring theme in these evaluations is a prevalent diversity collapse and ``artificial hivemind'' effect \citep{jiang2025hivemind} in which instruction-tuned models consistently use the same assistant-like language at the expense of diversity \citep{kirk2024understanding, padmakumar2024does}.
\citet{lu2026assistant} locate this collapse in an ``assistant axis'' within model activations that persists even with persona prompting.
\citet{yun2025price} show that the diversity collapse is reinforced by structured output templates that limit the output space during fine-tuning.
Base models, however, are not subject to these structural bounds. \citet{lin2026illusioninterventionllmsimulatedexperiment} shows that instruction-tuned models are sensitive to confounding in the presentation of queries, which is mitigated by using free-text negative confounders.
As \citet{lin2024urial} demonstrate, base models can achieve the same level of performance as instruction-tuned models in a broad, multi-faceted evaluation through prompting with stylistic examples while maintaining diversity.
We thus argue that the current reliance on instruction-tuned models is a misinformed default, and user simulation requires the breadth found in pretrained base models.

%% file: sections/s3_thesis.tex
\section{Pretrained and Instruction-Tuned LLMs for User Simulation}
\label{sec:thesis}
In this section, we formalize the problem of simulating human users in dialogue, where a language model is provided with a dialogue context and generates a continuation of the target human interlocutor's next turn. 
We explain how such generation is in itself a reflection of \textit{language modeling} capabilities of LLMs that is learned through pretraining on a large corpus of human text. 

\subsection{Two Classes of LLMs}
\label{sec:thesis:hlmm}
\textbf{Pretrained LLMs are Persona Mixture Models.}
At the utterance level, a pretrained language model defines a conditional distribution $p_{\theta}(u \mid c)$ over next-turn utterances $u$ given dialogue context $c$.  
We denote the corresponding true distribution of human utterances as $p^{\star}(u \mid c)$.
The training data for the model consists of many dialog samples from various speakers in particular states (emotions, beliefs, goals etc.)  $z$, but $z$ is not encoded explicitly in the data.
We can view this target distribution as a \emph{mixture} over implicit author states \(z\). 
Under this modeling lens,
\begin{equation}
\label{eq:mixture}
p^\star(u\mid c)
=
\int p^\star(u\mid c,z)\; p^\star(z\mid c)\,dz .
\end{equation}
Considered as a generative model for utterances, we can first {\em sample} $z$ using $p(z|c)$, i.e. choose a possible speaker, and then generate an utterance conditioned on $z,c$. Since the speaker is often weakly constrained by the context $c$, the distribution of $z$ is quite broad and the diversity of $u$ is similarly so (and matches the diversity of human dialog).

During pretraining, language models are trained under the cross-entropy loss over large-scale corpora, the vast majority of which are human-authored text. 
Training with this objective minimizes divergence of the model distribution $p_{\theta}$ from the empirical distribution over the training corpus, $\hat{p}^{\star}$:%
\begin{equation}
    \operatorname{arg\,min}_{\theta}\;
      \mathbb{E}_{(u,c)\sim \hat{p}^{\star}}
      \bigl[-\log p_{\theta}(u \mid c)\bigr].
\end{equation}
Minimizing this objective is equivalent to minimizing $\mathrm{KL}(\hat{p}^{\star} \| p_{\theta})$. 
Thus, in the idealized limit of scaled representative data and training convergence, the pretrained model distribution approaches the empirical conditional distribution of human-authored continuations. 
Assuming the corpus itself is representative, then $p_{\theta}(u\mid c) \approx \hat{p}^\star(u\mid c)\approx p^\star(u\mid c)$.

This view of cross-entropy training of LMs motivates our formulation  of pretrained LMs as Persona Mixture Models (PMMs), which we denote as $p_{\theta_{\mathrm{H}}}$.  
Following \citet{andreas2022language}, we adopt the view that  as an \emph{emergent} consequence of next-token prediction over a corpus produced by many distinct human authors, a pretrained LM maintains internal representations that correspond to latent states of a speaker.
In turn, this is precisely the property that user simulation requires. 
A PMM is trained to provide \emph{coverage} over the space of plausible human speakers. This in turn leads to outputs which provide the lexical, stylistic, and pragmatic variation present in natural human dialogue.


We note that an LLM context 
that specifies a few demographic traits (or
other user traits) does not specify a particular
user. Individual people have vastly more diversity than their
demographics, or any short trait description, and therefore any such description massively underspecifies an individual. One consequence of this is that if some short tuple of traits is used by an LLM to generate user dialog that appears to express more (new) traits in two independent runs, there will likely be inconsistencies between
the new traits. This is not a model error. 
The experimenter has simply forked two subsets (really distributions) of
humans specified by the initial set of traits,
according the new trait values. If the goal is to produce output consistent with the new traits, 
then the new traits should always be appended to the context before subsequent queries, i.e. they 
are not somehow embedded in the prior context string or LLM activations over it. A solution to ``freezing'' the model persona to 
(approximately) a single user is to use long, narrative backstories as the context~\citep{moon2024virtual}.

\textbf{Instruction fine-tuning to Assistant Language Models (ALMs).}
Instruction fine-tuning, on the other hand, transforms pretrained LLMs into helpful conversational assistants. A strong objective during post-training is {\em predictability and consistency}, that is, the model should produce substantially similar response to the same prompt, and it should use a consistent agent ``voice''.

In terms of Equation (\ref{eq:mixture}), the ``personality'' component of $z$ is largely independent of $c$, and $p(z|c)$ captures only non-persona traits like goals. 
While effective at aligning models with assistant-style desiderata, post-training removes the natural flexibility that pretrained models have to ``bind'' to a personality consistent with the context $c$.
The model is also more sensitive to particular words that it construes as direct or implicit instructions \citep{wu2026largelanguagemodelssensitive}.
\subsection{Tandem Models}
\label{sec:thesis:tandem}

As our experiments show and we have argued theoretically, pretrained models are generally much more accurate, but are
still prone to occasional out-of-domain responses
and decay of long-term human intentions. To address this,
recent works propose supervised fine-tuning a pretrained model directly on human conversational data, using objectives that remain anchored to the empirical distribution of human text~\citep{naous2026flipping}.
Other recent methods are inference-time or prompt-based techniques for extending the capabilities of pretrained models without weight updates.
\textit{Backstory conditioning}, for instance, prepends a rich narrative self-description of the simulated individual as model context, without requiring the model to interpret a prescriptive demographic profile as an instruction to enact as the user~\cite{moon2024virtual, kang2025deep}. 

These observations highlight a simple inference-time procedure, which we refer to as \emph{Tandem Modeling}.
This idea pairs a pretrained model as an utterance generator with an instruction-tuned model as a supervisor of the generated outputs from the utterance generator. 
First sample a pool of candidate utterances from the pretrained LLM; the supervising assistant model then selects a sampled candidate based on its selection/rejection criteria:
\begin{equation}
\begin{array}{rcl}
\mathcal{U}(c) &=& \bigl\{u^{(1)}, \ldots, u^{(N)}\bigr\},\quad u^{(i)} \sim p_{\theta_{\mathrm H}}(u \mid c), \\[6pt]
\hat{u} &=& \operatorname{arg\,max}_{u \in \mathcal{U}(c)} s_{\text{ALM}}(u, c).
\end{array}
\end{equation}
The key design constraint is that the IT model supervisor only selects or rejects, without altering or rewriting the final utterance: a simple approach to retain the generative diversity of pretrained models, while allowing the instruction-tuned model to bring its stronger discriminative capability to bear on which candidate best fits the local conversational context. 

\subsection{Experiment Setup}
\label{sec:thesis:exp_settings}
In the following sections, we provide experimental results supporting our claims on the differences between pretrained and instruction-tuned LMs in simulating human interlocutors.
To ensure that our findings are general across various domains of conversational language generation, we employ five conversational corpora spanning from open-domain dialogue to human-AI interactions. 
For each corpus we extract next-utterance prediction examples from multi-turn conversations (minimum one turn from each interlocutor) and generate $8$ continuations per context for each model. The total number of dialogue contexts we sample from the dataset are described in~\Cref{tab:corpora}.
Further details on the procedure and details of the text corpora are described in~\Cref{sec:appendix:experimental_details}.

We evaluate pairs of pretrained base LMs and the checkpoint after instruction-tuning across families of open-weight LLMs: Llama-3.1-8B, Mistral-7B, Mistral-Small-24B, Qwen2.5-7B, Qwen2.5-14B, Qwen3-8B, and Qwen3-14B. 
Comparing matched base/instruct pairs isolates the effect of instruction-tuning on user modeling.
Pretrained base models are prompted with the natural formatting of the dialogue so far; for instruction-tuned models, we provide the instructions ``You are simulating a user in the following dialogue'' and ``Continue the conversation as the user'' along with the string-formatted dialogue context.
For detailed examples and prompt formats, see~\Cref{sec:appendix:prompts}.

In the following sections, we mainly focus on model variants that are $>10$ billion parameters in size, and present full results in Appendix~\ref{sec:appendix:full_results}.
\input{tables/main/tab_corpora}

%% file: tables/main/tab_corpora.tex
\begin{table}[t]
\centering
\caption{\textbf{Conversational corpora used in our case study.} For all corpora, we use similar next-utterance prediction setups with LMs and sample  8 generations per context. For Multiwoz and DialOp, we take 128 and 117 dialogues, respectively, and generate utterances of all human-user turns; for all other datasets, we take the final user turn and generate model utterances in place of the human interlocutor. }
\small
\begin{tabularx}{\linewidth}{lXrr}
\toprule
Corpus & Dialogue type & Sampled Contexts & Total \\
\midrule
Reddit (ConvoKit)~\citep{chang2020convokit} & Open-domain; Human--Human & 1{,}024  & 8{,}192\\
WildChat-1M~\citep{zhao2024wildchat}& Human-LLM Chat; Human--AI & 1{,}024 & 8{,}192\\
LMSYS-Chat-1M~\citep{zheng2023lmsys}& Human-LLM Chat; Human--AI & 1{,}024 & 8{,}192\\
MultiWOZ 2.1~\citep{budzianowski2019multiwoz}& Task-Oriented; Human--Human & 807 & 6{,}456\\ 
DialOp~\citep{lin2024decision}& Task-Oriented; Human--Human & 306 & 2{,}448\\
\bottomrule
\end{tabularx}

\label{tab:corpora}
\end{table}

%% file: sections/s4_realism.tex
\section{Do LMs Estimate the Distribution of Human Utterances?}
\label{sec:realism}

In this section, we provide supporting analyses around our claim $\mathrm{C}$1 that pretrained LLMs are in fact models trained to estimate the (conditional) distribution of human language generation in a given dialogue context.
Conversely, we present evidence on how dominant post-training stages (RLHF/RLAIF and reasoning RL fine-tuning) alter model distributions from those learned during pretraining.


\subsection{Pretrained vs. Instruction-tuned LLMs through the Lens of Perplexity}
\label{sec:results:ppl}

A direct measure of how well a language model's induced conditional $p_\theta(u_t \mid c)$ aligns with the empirical distribution of human continuations $\hat{p}^{\star}(u_t \mid c)$ is its per-token perplexity on human dialogue text. 
Under the PMM view (Section~\ref{sec:thesis}), a faithful estimator of $p^{\star}$ should assign comparable likelihood to a human continuation and to its own samples drawn from $p_\theta$: both are draws from the same distribution the model is meant to capture. 

\input{figures/main/fig_ppl}
\textbf{Evidence for \texttt{C1}.}
If a language model successfully estimates the conditional distribution of human language given dialogue context, the perplexity on human continuations should be roughly similar to the perplexity on outputs sampled from the model. 

As shown in Figure~\ref{fig:ppl_reddit} and Table~\ref{tab:appendix_full_reddit}, instruction-tuned models assign very low perplexity to their own outputs but markedly higher perplexity to human text; pretrained models, on the other hand, show near-parity.
Comparing the same model before and after the post-training stages,  not only do  we revisit the known reports of mode-collapse from instruction-tuning in prior work, but also draw noteworthy findings on how pretrained base models assign high likelihoods of human-written continuations.

\subsection{Comparing the Human and Model-Generated Utterance Distributions}
\label{sec:results:mauve}

To complement per-token perplexity (a local, token-level probe), 
we then test global distributional similarity between model-generated and human-produced utterances using MAUVE~\citep{pillutla2021mauve}.
\input{figures/main/fig_mauve}

\textbf{The $\mathrm{MAUVE}$ metric.}
Let $\widehat P_{\mathcal D}$ and $\widehat Q_{\theta,\mathcal D}$  denote the empirical distributions over human-written and model-generated  utterances, encoded with a fixed text encoder $\phi(\cdot)$ and discretized into $K$ bins via $k$-means. 
For $\lambda \in (0,1)$, the mixture
$R_\lambda = \lambda \widehat P_{\mathcal D} + (1-\lambda)\,\widehat Q_{\theta,\mathcal D}$ admits two Kullback--Leibler
divergences: $\mathrm{KL}(\widehat P_{\mathcal D} \,\|\, R_\lambda)$ measures \emph{recall failure} (regions of human text the model fails to cover), and $\mathrm{KL}(\widehat Q_{\theta,\mathcal D} \,\|\, R_\lambda)$ measures \emph{precision failure} (regions where the model places mass humans rarely visit). 
Sweeping across
$\lambda$:
\begin{equation}
\mathcal{C}(\widehat P_{\mathcal D}, \widehat Q_{\theta,\mathcal D}) = \bigl\{\bigl(\exp(-c\,\mathrm{KL}(\widehat Q_{\theta,\mathcal D}\|R_\lambda)), \;
\exp(-c\,\mathrm{KL}(\widehat P_{\mathcal D}\|R_\lambda))\bigr) : \lambda \in (0,1)\bigr\}
\end{equation}

with scaling constant $c > 0$, $\mathrm{MAUVE}$ is the area under the divergence curve:
\begin{equation}
\label{eq:mauve}
\mathrm{MAUVE}(\widehat P_{\mathcal D},\widehat Q_{\theta,\mathcal D}) \;=\; \mathrm{AUC}\bigl(\mathcal{C}(\widehat P_{\mathcal D}, \widehat Q_{\theta,\mathcal D})\bigr)
\;\in\; [0, 1].
\end{equation}
A value of $1$ indicates the two samples are statistically indistinguishable; values below $1$ reflect mass placed off-support in either direction. 
MAUVE penalizes both failure modes symmetrically: a model cannot score highly by collapsing onto a high-confidence subset of human utterances, because that subset would leave most of $P$ uncovered.
In our experiments, we use a state-of-the-art embedding model (\texttt{gemini-embedding-2}) as the text encoder.

\textbf{Evidence for \texttt{C1}.} 
Figure~\ref{fig:mauve} shows MAUVE between model-generated and aligned human continuations for Reddit, LMSYS-Chat-1M, and WildChat-1M.
Pretrained base models sit near the ceiling ($0.95$--$0.97$) on all three corpora, consistent with the view that they estimate the distribution of human user text directly.
Instruction-tuned models drop substantially and unevenly; the Tandem configuration (see Section~\ref{sec:thesis:tandem}) shows near-identical scores as pretrained base models, showing that the distributional collapse is a property of direct generation with IT models.

%% file: figures/main/fig_ppl.tex
\begin{figure}[t]
\centering
\includegraphics[width=\linewidth]{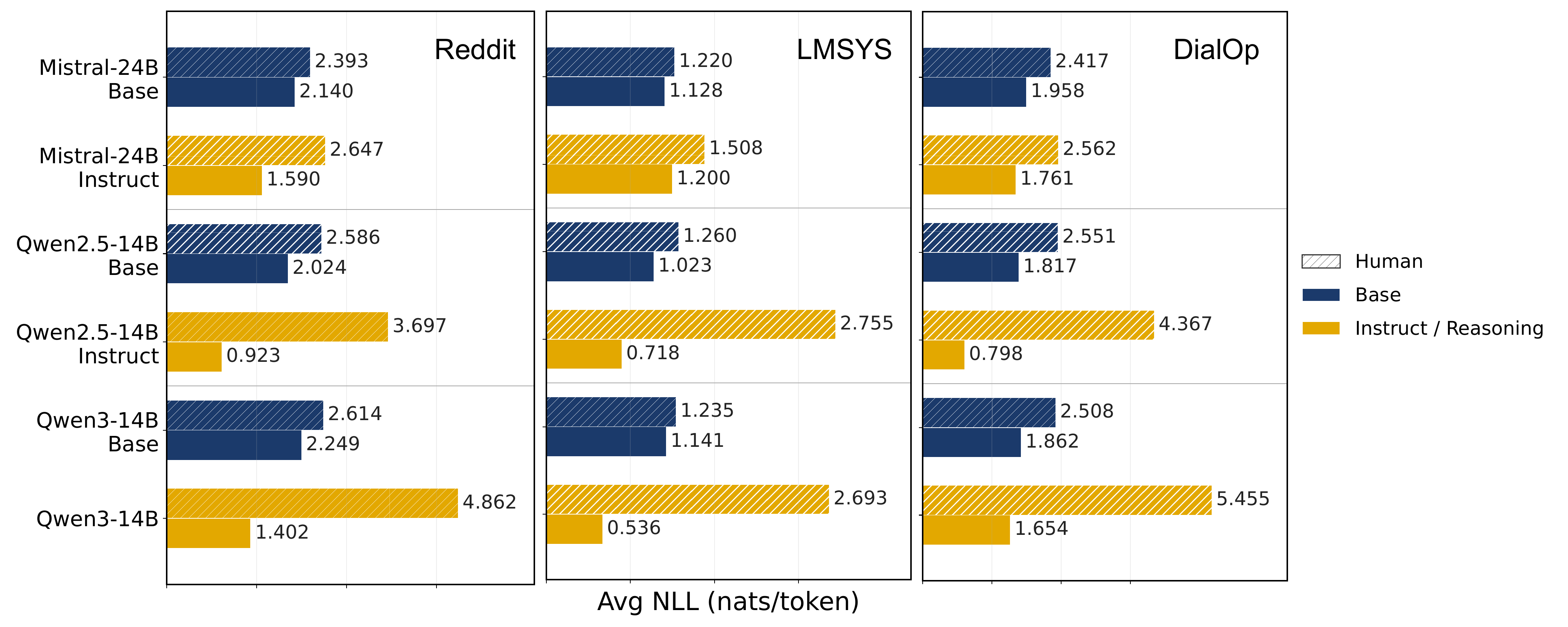}
\caption{\textbf{Per-token self-entropy and negative log-likelihood.}
For each model, we report model self-entropy (shown as solid bars) and negative log-likelihood of human utterances given dialogue context. See Appendix~\ref{sec:appendix:full_results} for full results.}
\label{fig:ppl_reddit}
\end{figure}

%% file: figures/main/fig_mauve.tex
\begin{figure}[t]
\centering
\includegraphics[width=\linewidth]{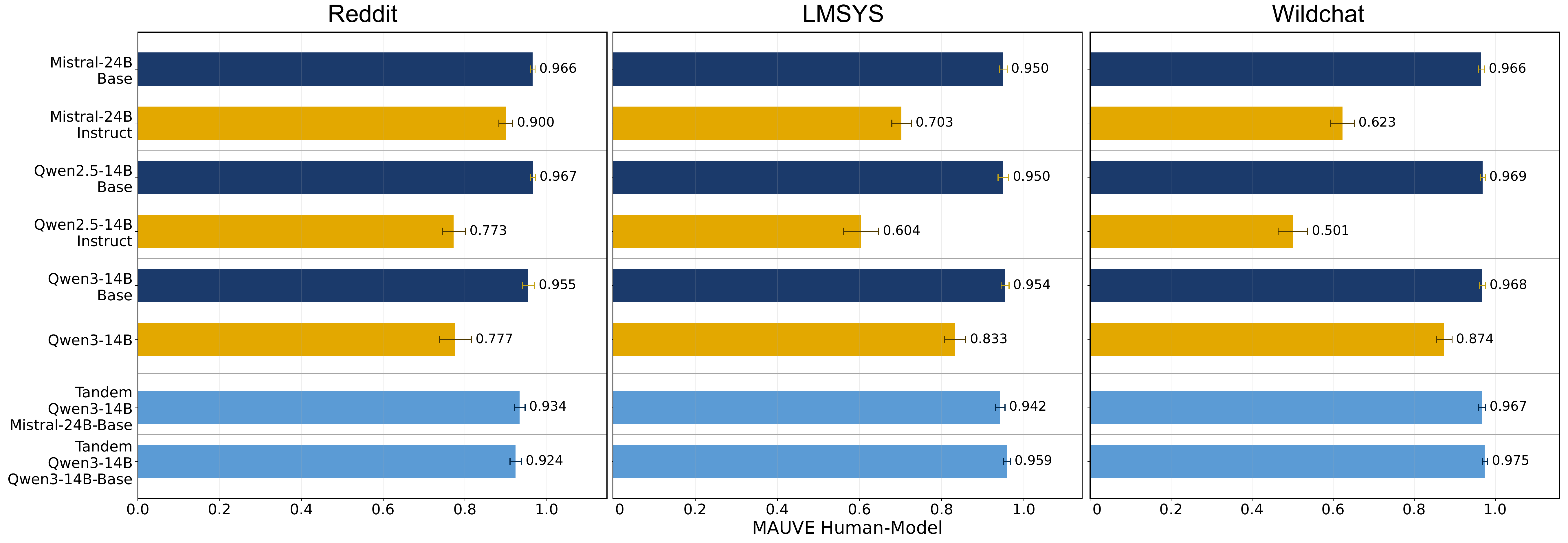}
\caption{\textbf{MAUVE results.} 
$\mathrm{MAUVE}$ is a sample-based score in $[0, 1]$ that quantifies how close two text distributions are. 
The score is high when the set of model samples and the set of human references cover the same regions of the distribution under a text embedding model; conversely, score is low when either distribution places mass where the other does not. 
A value near $1.0$ indicates that the two samples are statistically indistinguishable, while lower values mean the model is sampling from a different region of utterance space than humans. 
}
\label{fig:mauve}
\end{figure}

%% file: sections/s5_dialogue_acts.tex
\section{Do LMs Reflect Pragmatic Patterns of Human Interlocutors?}
\label{sec:dialogueacts}
Beyond utterance-level similarity, the realism of a user simulator in dialogue depends on its pragmatic consistency over multiple turns. This section provides analyses surrounding our claim \texttt{C2}: Pretrained LMs better reflect the pragmatic behaviors of human interlocutors.  

\textbf{Analysis on Distributions of Conversational Intents.}
Dialogue acts represent the communicative function of turns and their sequential organization \citep{stolcke2000dialogue}. To assess pragmatic competence, we build on prior evaluations of dialogue acts and the pragmatic competence of LLMs \citep{maitra2025dialogue, shaikh2024grounding, baidya2025behavior, shaikh2025navigating, yu2026pragmatic}. We compare model and human dialogue-act sequences on MultiWOZ-2.1 \citep{budzianowski2019multiwoz}, labeling generated user utterances with ConvLab BERTNLU \citep{henderson2020convlab} while using the aligned human turn annotations for the reference side. Each turn is converted to a speaker-marked intent token. Full details regarding labeling mechanics, intent inventory, and metric definitions are provided in Appendix~\ref{sec:appendix:dialogue_act_metric}.

\input{tables/main/tab_da_metrics}

We then compare human and model distributions over dialogue-act $N$-grams for $N=1,2, 3$. This sequence-level view extends utterance-level dialogue-act analyses of LLM behavior \citep{maitra2025dialogue, shaikh2024grounding}: a model may match marginal frequency of the dialogue act \texttt{Inform}, while still failing to reproduce when \texttt{Inform} follows a request or resolves a constraint. 

Table~\ref{tab:da_metrics_flat} shows Jensen-Shannon divergence (JSD) and vocabulary overlap with $k=100$ results for dialogue act sequences, and Figure~\ref{fig:da_analysis} visualizes the JSD between human and model distributions of dialogue acts across 1 to 3 turns. Results for models less than 10B as well as individual dialogue-act distributions can be found in Appendix~\ref{sec:appendix:full_results_multiwoz}.

\begin{table}
    \centering
    \small
    \captionof{table}{\textbf{Full JSD and Vocabulary Overlap results ($K=100$).} B, I, and R respectively indicate Base, Instruct, and Reasoning model variants.}
    \label{tab:da_metrics_flat}
    \vspace{5pt}
    \setlength{\tabcolsep}{2pt}
    \begin{tabularx}{\linewidth}{X ccc c ccc}
    \toprule
    & \multicolumn{3}{c}{JSD ($\downarrow$)} & & \multicolumn{3}{c}{V Overlap ($\uparrow$)} \\
    \cmidrule{2-4} \cmidrule{6-8}
    Model & $N=1$ & $2$ & $3$ & & $N=1$ & $2$ & $3$ \\
    \midrule
    Tandem & \textbf{.088} & \textbf{.219} & \textbf{.280} & & 0.08 & \textbf{.75} & \textbf{.84} \\
    \midrule
    M24B-B       & .181 & .277 & .330 & & 0.08 & \textbf{.75} & .74 \\
    M24B-I       & .195 & .299 & .346 & & 0.08 & .68 & .67 \\
    \midrule
    Q2.5-14B-B   & .175 & .272 & .325 & & 0.08 & .68 & .76 \\
    Q2.5-14B-I   & .119 & .273 & .321 & & 0.08 & .68 & .75 \\
    \midrule
    Q3-14B-B     & .180 & .282 & .331 & & 0.08 & .71 & .77 \\
    Q3-14B-R     & .184 & .354 & .389 & & 0.08 & .65 & .67 \\
    \bottomrule
    \end{tabularx}
\end{table}
\textbf{Evidence for \texttt{C2}.}  While all models diverge more from human patterns at higher n-gram orders, the Tandem approach maintains lower divergence with a JSD of 0.280 at $N=3$ compared to the 0.389 seen in the Qwen3-14B reasoning model. With the exception of Qwen2.5, the Instruct and reasoning models have higher divergence than their Base counterparts. The vocabulary overlap results support this; the Tandem retains the highest overlap through $N=3$, suggesting that picking from base model candidates preserves natural conversational transitions. Ultimately, the superior performance of the Tandem and Base models over the Instruct variants provides empirical evidence that pretrained LMs better reflect the pragmatic behaviors of human interlocutors by preserving natural transitions in task-oriented dialogue.
\input{figures/main/da_analysis_fig}

\textbf{Analysis on Distributional Preference Expression Rate.}
Preference expression rate refers to where in a dialogue each of the user preferences are expressed to the interlocutor. This section analyzes preference expression rate in DialOp's \citep{lin2024decision} travel-planning task. In this setting, a ``user'' expresses their preferences to an ``assistant'' in order to iteratively construct a travel itinerary. The location where a ``user'' expresses one of their assigned preferences is annotated using Gemini 2.5 Flash. For each utterance, our annotator receives the full list of ``user'' assigned preferences and a single ``user'' utterance from the dialogue. 
Our annotator is prompted to not infer any user intent and only label utterances if they are actively being expressed.
The prompt for our annotator is outlined in Section \ref{sec:appendix:DialOp_annotation}. 
To compare the preference expression rate of a model and human as a ``user'', we measure \textbf{(i)} the turn index $t \in {0, 1, 2, \dots}$ in which the annotated ``user'' utterance lies within a dialogue. And \textbf{(ii)} relative word position $w \in [0,100]$ of the middle word of an annotated preference, expressed as a percentage from 0 to 100 within the full sequence. 

\textbf{Evidence for \texttt{C2}.} Qualitative results from the annotator's outputs show that the difference in expression rate between base and instruction-tuned models is driven by how preferences are expressed, rather than how often. Base models tend to express their preferences via paraphrases of their initial preferences, whereas instruction-tuned models make explicit use of the wording in their preferences. This behavior inflates the preference-annotation count for instruction-tuned models vs. base models. Consider the opening turn of a DialOp conversation where the ``user'' has been assigned a private goal that includes: \emph{COVID conscious, outdoor seating places would be best} (among nine other items) and the assistant's first utterance is ``Hi, how are you doing today?'' The responses are as follows: 
\begin{quote}\small
\begin{tabularx}{\linewidth}{@{}lX@{}}
\textsc{\textbf{Human:}} &
``Great, I'm alright! Thanks.'' \hfill {\color{gray}\scriptsize user} \\

\textsc{\textbf{Base:}} &
``I am doing great. Thanks.'' \hfill {\color{gray}\scriptsize base model} \\

\textsc{\textbf{Instruct:}} &
``I'm good! I'm in town for the day and wanted to get some recommendations for places to visit.
Would be best if it's COVID friendly and places have outdoor seating.''
\hfill {\color{gray}\scriptsize instruction-tuned model}
\end{tabularx}
\end{quote}

                                      
The distributional measurements for $t$ and $w$ are normalized so the distributions are comparable, however the over-representation of preferences from the instruction-tuned model produces utterances that differ significantly from humans. Across model families, instruction-tuned simulators receive on average $3$--$7\times$ more preference annotations per utterance than either base simulators or human users (means of $2.8$--$6.6$ vs.\ $0.9$--$1.2$ vs.\ $0.9$ per utterance, respectively).


%% file: tables/main/tab_da_metrics.tex

%% file: figures/main/da_analysis_fig.tex
\begin{figure}[t]
\centering
\includegraphics[width=\linewidth]{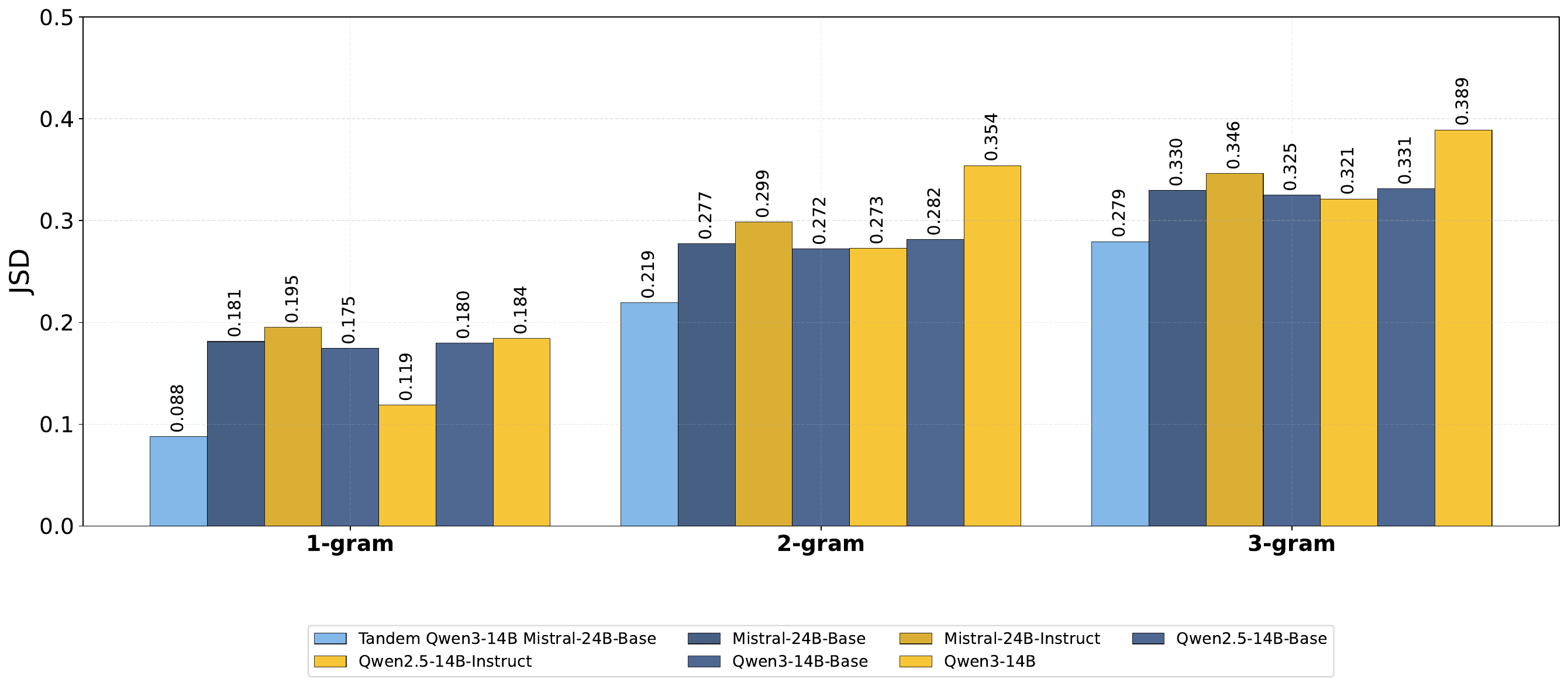}
\caption{
\textbf{Jensen--Shannon divergence (JSD) between human and model
dialogue-act $N$-gram distributions on MultiWOZ-2.1.} For each
$N \in \{1, \dots, 3\}$, we form the empirical distribution of
speaker-marked dialogue-act $N$-grams from the aligned human reference
($P_N$) and from the model under evaluation ($Q_N$), then plot
$\mathrm{JSD}(P_N, Q_N)$. JSD is bounded in $[0,1]$; \emph{lower is
better}, with $0$ corresponding to identical distributions. Higher $N$
probes longer pragmatic context---a model that matches marginal act
frequencies (low JSD at $N{=}1$) can still fail to reproduce the
\emph{ordering} of acts across turns (high JSD at $N \geq 3$). 
}
\label{fig:da_analysis}
\end{figure}

%% file: sections/s6_diversity.tex
\section{Do LMs Capture the Diversity of Language Generated by Human Speakers?}
\label{sec:diversity}
\input{figures/main/fig_diversity}

This section provides experimental evidence for our claim
\texttt{C3}: pretrained base models, viewed as Persona Mixture Models, capture the diversity of human speakers in a way that instruction-tuned assistant language models systematically fail to. 

\textbf{Metrics of Lexical and Semantic Diversity of Model Generations.}
We measure generation diversity at two complementary levels: lexical (Distinct-2, Self-BLEU~\cite{papineni2002bleu}, Self-ROUGE-L\cite{lin2004rouge}) and semantic variability (Self-BERTScore-F1\cite{zhang2020bertscore}, Vendi\cite{friedman2023vendi}) of model generations conditioned on the same input context.
Further details on the metrics and the experimental configurations are found in Appendix~\ref{sec:appendix:diversity_metrics}.

\textbf{Evidence for \texttt{C3}.}
Figure~\ref{fig:diversity} summarizes utterance-level diversity results on Reddit.
Across all three metrics---Self-BLEU, Self-BERTScore-F1, and Vendi score---pretrained base models generate substantially more varied continuations than their instruction-tuned or reasoning-tuned counterparts.
Lower Self-BLEU and Self-BERTScore-F1 values, which indicate greater lexical and semantic diversity respectively, are consistently observed for base models across every base/instruct pair. 
Vendi scores, where higher values indicate a greater effective number of semantically distinct samples, further confirm that instruction-tuning leads to a marked collapse in output diversity.
Notably, the Tandem configuration preserves near-base diversity while still benefiting from IT model-based candidate selection.

%% file: figures/main/fig_diversity.tex
\begin{figure}[t]
\centering
\includegraphics[width=\linewidth]{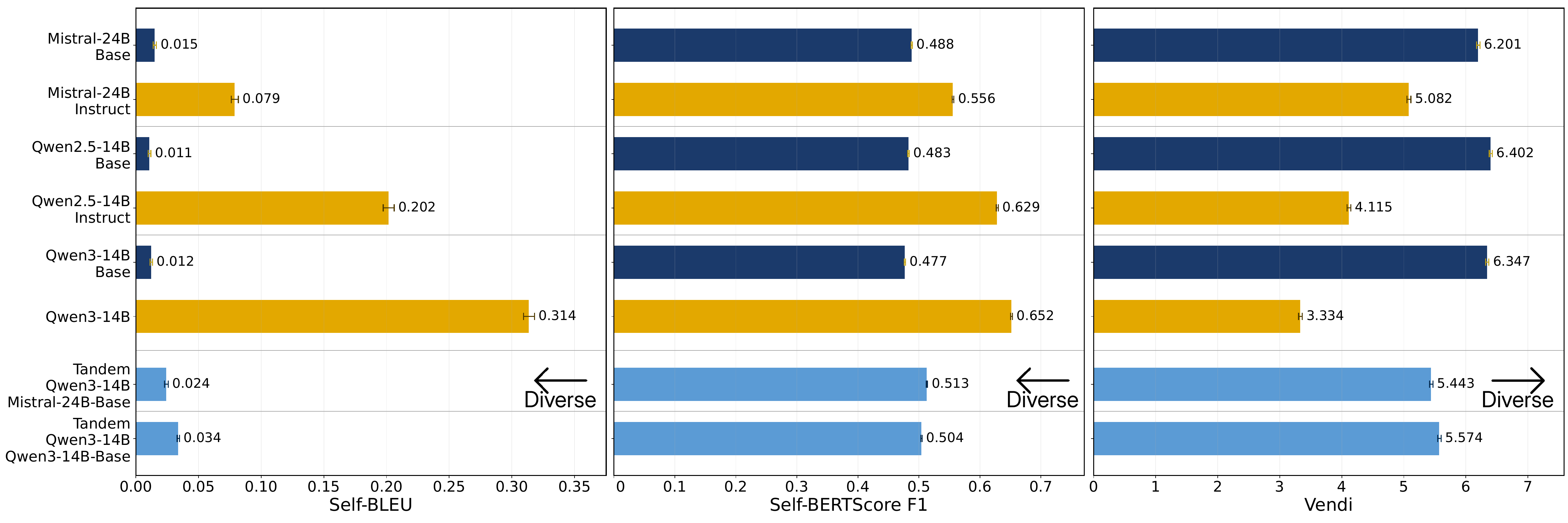}
\caption{
\textbf{Utterance-level diversity diagnostics on Reddit.} For each model we report three complementary measures of how varied the model-generated continuations are within a given dialogue context, averaged across contexts. 
\textbf{Self-BLEU} (left) computes BLEU between every pair of continuations from the same context: \emph{lower is better}, since a low value means the continuations are lexically distinct from one another. \textbf{Self-BERT-F1} (center) is the analogous quantity in contextual-embedding space, capturing semantic rather than surface variation: again, \emph{lower indicates greater diversity}. 
\textbf{Vendi score} (right) reports the effective number of semantically distinct samples, computed from the eigenspectrum of the embedding similarity kernel~\citep{friedman2023vendi}: \emph{higher is better}, with a value near $1$ corresponding to total mode collapse.
}
\label{fig:diversity}
\end{figure}

%% file: sections/s7_challenges.tex
\section{Challenges for LLM Human Simulation}
\label{sec:challenges}
Here we discuss some outstanding challenges for PMM-based human simulation. 

\subsection{How to test Human-ness: {\em What} would Turing do, and {\em How} would he do it?} As suggested by our perplexity plots earlier, PMMs bound to personas by dialog histories produce text that is extremely close to the original human text. To reduce the gaps between real and simulated responses further, its likely that feedback will be needed from human/sim classifiers as in GANs (Generative Adversarial Networks). For instance, an LLM real/sim classifier model can be used to provide a reward signal for a generator LLM via RL training~\citep{yu2024finetuninglanguagemodelsgenerative}.  But what kind of classifier, and what kind of data to train it on? This matters because we are trying to distinguish human behavior in all its complexity from
mechanical simulacra of it. In this setting, ``what'' is being said (some assertions about the world state) is arguably less important than ``how'' it is being said (which encodes all the inner complexity of the
speaker's emotions, beliefs, hopes, conflicts, yearnings etc). Language models are potentially capable of modeling much of this and are a natural
substrate for such classifiers, but we caution 
against the use of ``semantic embeddings'' as input to a human/sim classifier. ``Semantics'' as commonly
understood focuses on an idealized ``what'' about the world state. Such embeddings typically ignore the emotional, cultural, and psychological state encoded in an utterance. Using an LLM's full internal state (or that of some of its layers) rather than a semantic embedding, should be preferable as input to classifiers for human/sim classification. Finally, such a tuning dataset is likely to be very large, since a pretrained model has already been trained to minimize cross-entropy loss (i.e. distributional distance) on an enormous dataset. 

In terms of the practical application of GAN-tuning
of human simulation models, an important use case is simulation of human users during
chat model optimization. If the chat model is able
to distinguish real from synthetic users from signals
in its activations, then it may ``game'' the training and apply its reinforcement signals only to that subset. i.e.  it may learn only to improve its performance on synthetic users and not real users. GAN optimization using a discriminator based on the full latent state of the agent model (or similar LLM) makes the human/synth distinction invisible to the agent model, and should guard against such gaming. 

A problem with using naive classifiers (including human judges) for human/sim classification is that they have difficulty taking diversity into
account. In~\citep{jones2025turing} it was asserted that LLMs ``Pass the Turing Test'' because both LLMs
and human judges consider the responses of certain
LLMs to be more human-like than actual human responses in pairwise comparisons. In one case, a GPT model achieved a success
rate of 70\% (the judge classified the synth result as human 70\% of the time). However, we question whether Turing would accept that result. If the goal is solely to fool human judges, then the claim is fair. But if
the goal is to achieve a distribution of simulated responses
that is truly realistic (statistically indistinguishable
from human responses), then we are still some distance
from that goal. The classifiers performance of 70\% shows that it was distinguishing synth and human quite well, albeit while applying the wrong label. 
It is less impressive if one recognizes that humans (and LLM judges) are easily fooled by
certain artifacts in responses, namely their lack of
diversity. If we assume the human and machine judges
score various responses as more- or less- human-like, then a model attempting to fool them should always output the most human-like response, with no diversity. A real human will produce a variety or responses, but will score lower than a synth that
always produces the preferred response. The solution is to train a classifier which works on two {\em distributions} of responses (real and simulated) and is updated after each update to the generators. Any lack of diversity in the synthetic responses will make
them easy to distinguish. The GAN configuration we described earlier achieves this. 

\subsection{Nescience: Its not what you know but what you {\em Don't} that makes you human.}

One of the hardest characteristics of humans to reproduce with LLMs simulation is age/education-appropriate level of knowledge. Both pretrained LLMs
(which include Wikipedia and open-source books in their training data) and IT LLMs (which include far more specialized knowledge) struggle to ``unknow'' what they know based on a persona binding. Thus it is difficult to get them to reproduce the difficulties
that less-knowledgeable users have on a variety of tasks. Similarly, it can be hard for LLMs to try to solve a problem using a (realistically human) wrong method. This is particularly relevant to the use of LLMs in educational contexts to model and remediate
student misconceptions.

On the positive side, pretrained LLMs  can reproduce various
false or incomplete facts, and erroneous solution
methods, that are typical of real humans. The rate of these errors is usually lower than for real humans, but many typical errors are represented. We are exploring a more elaborate use of tandem models, 
where a supervisor model adjudicates wether the
base model should or should not be able to use a
given piece of information or solution method. If the supervisor model is able to estimate the {\em probability} that a user with given background would
know a fact or method, it can accept their response
with that probability.
If not, it can cause the PT model to repeat the request but with an additional instruction not
to use a particular fact or method. 

\subsection{The Limits of LLM training data.}
To predict what kinds of behavior can or cannot
be simulated by LLMs, its useful to refer back to
the genres of training data available to them. 
Most of the dialog data is relatively short-duration
or asynchronous dialog. It is rare to see longer-term
data, but there are exceptions:

\paragraph{The passage of time.}
People sometimes journal or keep diary entries
on various sites. This kind of data incorporates changes over months or years. We are exploring the
simulation of behavior change in humans using
simulated daily diary entries.

\paragraph{Acting vs Saying.}

Language models record utterances or passages of text by various authors in different contexts. There is no direct recording of the authors' actions; however, people do spend
a lot of time talking about what they have done. 
They also spend quite a lot of time talking about what
they want or intend to do. However, there is a world
of difference between the two. Asking a human what they did in most cases yields a straightforward account of what they did. Asking what a user {\em will do} asks for an expression of intention, or
perhaps desire, or perhaps goal. In many cases, this
intention is not realized, or the action taken is far from the intention. So asking for action choices in future is fraught with error. Asking
what a virtual person {\em just did} is much more
reliable. The questioner can easily move virtual time
until just after the action happened to afford this.

\paragraph{First, second and third order statistics.} Estimates of high-order statistics
of words modeled by LLMs should be intuitively 
more difficult as the order increases. Examples 
of first-order traits are basic frequencies of 
words which vary by language, accent, education,
and age. These are extremely well-represented in the training data. So should be the dominant principal components of word frequencies, such as basic personality types which are known to be encoded in subject's lexicons (aka the lexical hypothesis of personality).

However, virtual subjects personalities are often
assessed using surveys such as IPIP50 and BFI which involves self-reporting. Virtual subjects' ability
to answer faithfully depends on second-order training
data, i.e. examples of text that manifest the
subject's personality pattern in their lexicon, plus
a self-description term. Since the lexical patterns are ubiquitous and the self descriptors use common words, its not surprising that such interview-based personality assessments match well with human norms. But personality trait correlations based on those interviews are higher than human norms.
This measurement involves {\em third-order} patterns
in the training data, i.e. the personality lexical 
dimensions plus two personality self-description terms. The sheer number of such personality 
descriptor combinations is also higher, making accurate estimation more difficult.



%% file: sections/s8_conclusion.tex
\section{Conclusion and Limitations}
\label{sec:conclusion}

In this work, we have argued that user simulation is a very different task from dialog assistance. The optimizations applied for the assistant task {\em systematically degrade} performance for
human simulation, and IT models are 
as well-adapted for human simulation as
unaligned pretrained models are for the assistant task. We adopted the terms Persona Mixture Models (PMMs) to better capture the distinction between these models (currently realized as pretrained models) and ALMs (IT assistant models).

\textbf{Limitations.} Our case study focused on five English-language corpora; extensions of our analysis to multilingual settings, domain-specific dialogue, and/or synchronous spoken interaction remain as open empirical questions.
No single metric we report is, on its own, a definitive measure of user simulation fidelity, and such a definition remains contested in the literature. 
Our experimental support rests instead on the convergence of independent diagnostics: perplexity, MAUVE, dialogue-act structure, and lexical and semantic diversity all point in the same direction, across five corpora and seven model families. We invite future work to expand, affirm, or challenge these methodologies.

We point out that base model user simulators can produce outputs that might be considered harmful, which instruction-tuning and RLHF are intended to reduce. Our target applications normally direct PMM output to a human study or to an agent LLM, where the presence of that content may be an important aspect of realism. For example, the production of harmful content allows for realistic community simulations, for developing of mechanisms of moderation. While these applications do not expose human peers to PMM output,  future applications, e.g. hybrid human/sim community simulations, may do so. In those cases, the standard protections for human-produced content can be applied, and in addition the supervisor LLM in a tandem model can be instructed to filter out harmful content, or to enforce community standards. 

LLM pretraining corpora  are themselves large, poorly characterized, and subject to their own distributional biases~\cite{bender2021dangers}. 
The user models that emerge from pretraining reflect the demographics and discourse norms captured in web-scale data. 
IT models typically introduce additional biases from post-training data which are
based on behavioral norms adopted by their designers. These may or may not represent the full user community.
A natural follow-on question to our work is which user populations are over- or under-represented in the implicit speaker mixture of LLMs after pretraining, and whether targeted data curation, lightweight fine-tuning or adaptation of an interviewer ``voice'' can address those gaps without reintroducing the diversity collapse that instruction-tuning produces.

\textbf{Broader implications.}
The broader implication of this work is that user modeling deserves a dedicated methodology, sensitive to the target dialogue regime and distinct from the objectives, failure modes, and evaluation criteria of assistant training. 
Solutions developed for assistant alignment do not automatically transfer to user simulation; in many cases, they actively work against it. We hope that the framework we have laid out here, including the PMM/ALM distinction, the three simulation desiderata C1--C3, and the example Tandem modeling idea, provides a useful conceptual scaffold for the community, and that the experimental evidence encourages more careful treatment of model selection when LLMs are deployed as proxies for human users.

%% file: sections/a2_experimental_details.tex
\section{Experimental Details}
\label{sec:appendix:experimental_details}

\subsection{Next-Utterance Prediction Protocol}
\label{sec:appendix:next_utterance_protocol}

All experiments instantiate user simulation as next-utterance prediction. Given a
dialogue prefix $c=(x_1,\ldots,x_{t-1})$ and target user $u$, a simulator
generates a candidate continuation $\hat{x}_t$ for the same speaker and turn at
which the human continuation $x_t$ was observed. This setup keeps the discourse
state, partner behavior, and target speaker fixed while varying only the
simulator's next user utterance.

For Reddit, WildChat-1M, and LMSYS-Chat-1M, we sample target user turns from
multi-turn conversations after the opening turns. For task-oriented corpora, the
evaluation expands conversations into user-turn prediction examples so that
later-turn pragmatic structure is represented. Each retained context is paired
with the aligned human continuation and with multiple simulator continuations.
Unless otherwise stated, each model produces eight continuations per context.

\subsection{Corpus Preparation}
\label{sec:appendix:corpus_preparation}

\textbf{Reddit.}
The Reddit experiments use the ConvoKit Reddit corpus. Threads are rendered with
the thread title followed by speaker-labeled comments. Deleted or removed turns
are excluded. The sampled target turn must belong to a two-participant exchange,
and consecutive runs from the same speaker are not used as new target turns. The
resulting task is open-domain human-human continuation in informal online
discussion.

\textbf{WildChat-1M.}
WildChat-1M conversations are filtered to English, non-toxic, non-duplicate
multi-turn human-assistant conversations. Target turns are restricted to human
user turns. Contexts are rendered as alternating \texttt{User:} and
\texttt{Assistant:} turns, and the target is the next user utterance.

\textbf{LMSYS-Chat-1M.}
LMSYS-Chat-1M is treated as a second real-world human-LLM chat corpus. English
multi-turn conversations are retained, redacted conversations are dropped, and
exact duplicate records are removed. As in WildChat-1M, target turns are user
turns and contexts are rendered with \texttt{User:} and \texttt{Assistant:}
labels.

\textbf{DialOp.}
DialOp is converted from itinerary-planning action logs into alternating
user-assistant messages. Player 0 is treated as the user and player 1 as the
assistant. Private itinerary goals and preferences are preserved when available
and are shown to the simulator as user-private information. The target is a user
utterance in a goal-directed planning dialogue.

\textbf{MultiWOZ.}
MultiWOZ 2.1 is used for task-oriented dialogue-act analysis. We use labeled
dialogues with turn-level dialogue-act annotations. User goals are retained in
the context when available, and target user turns are expanded into examples for
generation and subsequent dialogue-act labeling.

\subsection{Model Families}
\label{sec:appendix:model_families}

We evaluate three simulator families. Direct PMM systems are base/pretrained
language models sampled directly from the dialogue context. Direct ALM systems
are instruction-tuned variants prompted to continue the conversation as the
human user. Tandem systems use a PMM as a candidate speaker and an ALM as a
selector. Matched base/instruct comparisons isolate the effect of instruction
tuning within a model family.

\subsection{Direct Generation}
\label{sec:appendix:direct_generation}

Base models receive a plain-text completion prompt consisting of the rendered
dialogue context followed by the target speaker label. Generation stops at
dataset-specific strings that mark the beginning of a new turn, such as a later
\texttt{User:}, \texttt{Assistant:}, Reddit username marker, or thread boundary.
Instruction-tuned models receive a system/user chat prompt asking them to
continue as the user and to output only the next user utterance.

Generated text is normalized before evaluation. Leading speaker labels are
removed, generated reasoning blocks are stripped when present, and outputs that
continue into later role-labeled turns are truncated to the first utterance.
Empty generations are discarded.

\subsection{Tandem Generation}
\label{sec:appendix:tandem_generation}

The tandem simulator separates candidate production from candidate selection.
For a dialogue context $c$, the PMM speaker samples a candidate set
$\mathcal{U}(c)=\{u^{(1)},\ldots,u^{(N)}\}$. The ALM selector reads the same
dialogue context and the candidate set, reasons about communicative fit, and
selects one candidate:
\[
  \hat{u} = \arg\max_{u \in \mathcal{U}(c)} s_{\mathrm{ALM}}(u,c).
\]
The selector is constrained to return a candidate verbatim. This prevents the
ALM from rewriting the surface form into its own assistant-like style while
still allowing deliberative selection for local relevance and consistency.

We evaluate two tandem variants. The \emph{User Agent} variant first produces an
explicit belief-desire-intention plan for the next turn, then scores candidates,
then emits the selected candidate verbatim. The \emph{Critique Only} variant
omits the explicit planning step and selects the first candidate that satisfies
all rubric criteria.

\subsection{Sampling and Splits}
\label{sec:appendix:splits}

Generation examples are shuffled with a fixed random seed before sampling. The
default evaluation sample contains 1,024 contexts for Reddit, WildChat-1M, and
LMSYS-Chat-1M; task-oriented corpora are expanded from sampled dialogues into
their eligible user turns.

\subsection{Output Validation}
\label{sec:appendix:output_validation}

All evaluation rows require a nonempty simulator continuation and an aligned
human continuation. Diversity metrics are computed after text preprocessing and
per-context deduplication.

%% file: sections/a3_prompts.tex
\section{Prompt Templates}
\label{sec:appendix:prompts}

This appendix gives the prompt templates used for generation and tandem
selection. Braced fields denote dataset-specific values inserted at evaluation
time.

\subsection{Direct User-Simulation Prompts}
\label{sec:appendix:direct_prompts}

\begin{tcblisting}{enhanced,breakable,colback=sub,colframe=main,boxrule=0.6pt,arc=2pt,title={Reddit instruction-tuned simulator: system prompt},listing only,listing options={basicstyle=\ttfamily\footnotesize,breaklines=true,columns=fullflexible,keepspaces=true}}
You are simulating a Reddit user engaging with conversations with different users. You are to simulate the user u/{user_id}. You will be given past conversations that this user has engaged with, and based on that, continue the conversation naturally as the user. Only answer with the next user utterance and no other text.
\end{tcblisting}

\begin{tcblisting}{enhanced,breakable,colback=sub,colframe=main,boxrule=0.6pt,arc=2pt,title={Reddit instruction-tuned simulator: user prompt},listing only,listing options={basicstyle=\ttfamily\footnotesize,breaklines=true,columns=fullflexible,keepspaces=true}}
Here is the conversation context, consisting of Reddit thread(s):

###THREADS START

{context}

###THREADS END

Generate a comment that is in line with the final thread conversation for user u/{user_id}. Continue the conversation naturally as the user.Only answer with the comment and no other text.
\end{tcblisting}

\begin{tcblisting}{enhanced,breakable,colback=sub,colframe=main,boxrule=0.6pt,arc=2pt,title={Human-LLM chat instruction-tuned simulator},listing only,listing options={basicstyle=\ttfamily\footnotesize,breaklines=true,columns=fullflexible,keepspaces=true}}
System:
Continue the conversation naturally as the user. Only answer with the next user utterance and no other text.

User:
{context}

Continue the conversation naturally as the user. Only answer with the next user utterance and no other text.
\end{tcblisting}

\begin{tcblisting}{enhanced,breakable,colback=sub,colframe=main,boxrule=0.6pt,arc=2pt,title={DialOp instruction-tuned simulator},listing only,listing options={basicstyle=\ttfamily\footnotesize,breaklines=true,columns=fullflexible,keepspaces=true}}
System:
You are simulating a user engaging with conversations with a different user on planning a travel itinerary. You are to simulate the user marked as 'User'. You will be given past conversations that this user has engaged with, and based on that, you must generate a comment that is like the given user.

User:
{goal_block}
Here is the conversation context so far:

###CONVERSATION START

{context}

###CONVERSATION END

Generate an utterance that is in line with the final thread conversation for the User. Only answer with the utterance and no other text.
\end{tcblisting}

\begin{tcblisting}{enhanced,breakable,colback=sub,colframe=main,boxrule=0.6pt,arc=2pt,title={MultiWOZ instruction-tuned simulator},listing only,listing options={basicstyle=\ttfamily\footnotesize,breaklines=true,columns=fullflexible,keepspaces=true}}
System:
You are a simulating a user engaging in a task-oriented dialogue. Continue the conversation as the USER. Only answer with the next USER utterance and no other text.

User:
Dialogue context:
{context}

Continue the conversation as the USER. Only answer with the next USER utterance and no other text.
\end{tcblisting}

\subsection{Pretrained Base-Model Prompts}
\label{sec:appendix:base_prompts}

Pretrained base models were prompted as continuation models rather than as
instruction-following chat assistants. The prompt is the serialized dialogue
context itself, with no system message or natural-language instruction appended.
Generation begins immediately after the final target-speaker prefix. The same
completion-style prompt is used for direct base-model generation and for the
base utterance generator inside tandem runs.

\begin{tcblisting}{enhanced,breakable,colback=sub,colframe=main,boxrule=0.6pt,arc=2pt,title={Reddit pretrained base simulator},listing only,listing options={basicstyle=\ttfamily\footnotesize,breaklines=true,columns=fullflexible,keepspaces=true}}
[Title]: {thread_title}

[Original Poster (OP) u/{op_user_id}]: {op_text}

[u/{speaker_id}]: {utterance}

...

[u/{target_user_id}]:
\end{tcblisting}

When the target speaker is the original poster, the final Reddit prefix uses
the same original-poster form as the context, i.e.,
\texttt{[Original Poster (OP) u/\{target\_user\_id\}]:}.

\begin{tcblisting}{enhanced,breakable,colback=sub,colframe=main,boxrule=0.6pt,arc=2pt,title={Human-LLM chat pretrained base simulator},listing only,listing options={basicstyle=\ttfamily\footnotesize,breaklines=true,columns=fullflexible,keepspaces=true}}
User: {user_utterance}

Assistant: {assistant_utterance}

...

User:
\end{tcblisting}

This continuation template is used for both WildChat-1M and LMSYS-Chat-1M.

\begin{tcblisting}{enhanced,breakable,colback=sub,colframe=main,boxrule=0.6pt,arc=2pt,title={DialOp pretrained base simulator},listing only,listing options={basicstyle=\ttfamily\footnotesize,breaklines=true,columns=fullflexible,keepspaces=true}}
User goals:
{goal_bullets}

Conversation so far:

User: {user_utterance}

Assistant: {assistant_utterance}

...

User:
\end{tcblisting}

\begin{tcblisting}{enhanced,breakable,colback=sub,colframe=main,boxrule=0.6pt,arc=2pt,title={MultiWOZ pretrained base simulator},listing only,listing options={basicstyle=\ttfamily\footnotesize,breaklines=true,columns=fullflexible,keepspaces=true}}
USER GOAL: {goal}

--------------------

USER: {user_utterance}

SYSTEM: {system_utterance}

...

USER:
\end{tcblisting}

\subsection{Tandem User-Agent Prompts}
\label{sec:appendix:tandem_prompts}

\begin{tcblisting}{enhanced,breakable,colback=sub,colframe=main,boxrule=0.6pt,arc=2pt,title={User Agent system prompt},listing only,listing options={basicstyle=\ttfamily\footnotesize,breaklines=true,columns=fullflexible,keepspaces=true}}
You are simulating how a human would respond. Your task is to generate authentic, human-like replies that are consistent with the user description below.

## User description
{user_description}

---
## Your process for each turn

Each time you need to produce a reply, follow exactly two steps.

### Step 1 -- Plan and request candidates
### Step 2 -- Select appropriate candidate
\end{tcblisting}

\begin{tcblisting}{enhanced,breakable,colback=sub,colframe=main,boxrule=0.6pt,arc=2pt,title={User Agent planning prompt},listing only,listing options={basicstyle=\ttfamily\footnotesize,breaklines=true,columns=fullflexible,keepspaces=true}}
### Step 1 -- Plan the next turn

Open a <user_think> block. Inside it:
1. **Belief**: Summarise what has happened in the conversation and what you currently understand.
2. **Desire**: State what you want to achieve in this turn (your conversational goal).
3. **Intention**: State the specific communicative act you will perform (e.g. "share an anecdote", "ask a clarifying question", "express mild disagreement").

Do not produce the final reply yet. Candidate replies will be provided in the next step.

Close the </user_think> block.
\end{tcblisting}

\begin{tcblisting}{enhanced,breakable,colback=sub,colframe=main,boxrule=0.6pt,arc=2pt,title={User Agent candidate-selection prompt},listing only,listing options={basicstyle=\ttfamily\footnotesize,breaklines=true,columns=fullflexible,keepspaces=true}}
### Step 2 -- Analyze candidates

You have receive a numbered list of candidate utterances.
{tool_result_str}

Open a new <user_think> block.
Score each candidate on the following rubric:

{rubric_text}

Write a brief score (1-10) and one-sentence reason for each candidate, then name the best.
Close the </user_think> block.

Do not copy the final candidate utterance in this step. The next step will ask for the final exact candidate text.
\end{tcblisting}

\begin{tcblisting}{enhanced,breakable,colback=sub,colframe=main,boxrule=0.6pt,arc=2pt,title={User Agent final-selection prompt},listing only,listing options={basicstyle=\ttfamily\footnotesize,breaklines=true,columns=fullflexible,keepspaces=true}}
### Step 3 -- Final candidate text

Using your candidate analysis above, select the best candidate from the numbered list. Output only the EXACT text of that one candidate.

Do NOT include reasoning, markdown, candidate numbers, labels, quotes, or XML tags.
Do NOT modify, rephrase, shorten, or add anything.
Your entire response must be character-for-character identical to one of the numbered candidates.

You MUST select one candidate.
\end{tcblisting}

\begin{tcblisting}{enhanced,breakable,colback=sub,colframe=main,boxrule=0.6pt,arc=2pt,title={Critique Only tandem prompt},listing only,listing options={basicstyle=\ttfamily\footnotesize,breaklines=true,columns=fullflexible,keepspaces=true}}
You are simulating how a human would respond.
Candidate utterances will be sampled from another model. Your task is to select the first candidate reply that meets all criteria in the rubric.

## User description
{user_description}

Do not rewrite candidates. Do not produce a new reply. Return only an exact candidate utterance after your critique.

### Candidate critique and selection

You have received a numbered list of candidate utterances:
{tool_result_str}

Evaluate candidates in ascending order against every criterion in this rubric:

{rubric_text}

Open a <user_think> block.
For each candidate, briefly state PASS or FAIL for each criterion. Then select the FIRST candidate that passes ALL criteria.
When multiple candidates meet all criteria, select the first passing candidate in the list.

Close the </user_think> block.

After </user_think>, copy the EXACT text of the chosen candidate -- do NOT modify, rephrase, shorten, or add anything. Your output after </user_think> must be character-for-character identical to one of the numbered candidates.

You MUST select one of the candidates. Do NOT reject all candidates or write in your own text outside of the provided options.
\end{tcblisting}

\subsection{DialOp Annotation Prompts}
\label{sec:appendix:DialOp_annotation}

\begin{tcblisting}{enhanced,breakable,colback=sub,colframe=main,boxrule=0.6pt,arc=2pt,title={DialOp 
  preference annotator: system prompt},listing only,listing                                           
  options={basicstyle=\ttfamily\footnotesize,breaklines=true,columns=fullflexible,keepspaces=true}}   
  You are an expert annotator for a trip-planning dialogue dataset.                                   
                                                                           
  Player 0 (the "user") was assigned a list of preferences before the conversation. Player 1 
  (the "agent") helps plan an itinerary. Your job is to determine whether a given Player 0 utterance  
  expresses any of their assigned preferences.                                                        
                                                   
  Rules:                                                                                              
  - Only identify preferences that Player 0 is actively expressing in this utterance. Do not guess or 
  infer preferences that are not clearly present.
  - For each preference you identify and extract the specific words/phrase from the utterance that
  express it. Do not return the entire utterance -- only the relevant substring(s). The extracted text
   must appear verbatim in the utterance.                                                             
  - A single utterance can express zero, one, or multiple preferences.
  - If the utterance does not express any preference (e.g. "Yes", "Greetings!"), return an empty list.                                             
  \end{tcblisting}    

  \begin{tcblisting}{enhanced,breakable,colback=sub,colframe=main,boxrule=0.6pt,arc=2pt,title={DialOp 
  preference annotator: user prompt},listing only,listing                                             
  options={basicstyle=\ttfamily\footnotesize,breaklines=true,columns=fullflexible,keepspaces=true}}   
  ## Player 0's Assigned Preferences                                                                  
  {preferences}
                                                                           
  ## Player 0's Utterance                     
  "{utterance}"                           

  ## Task                                                                                             
  Identify which (if any) of the assigned preferences Player 0 is expressing in this utterance. For
  each match, extract the verbatim text from the utterance that conveys the preference.               
                                          
  Return a JSON list:
  [                                                                                                   
    {
      "preference_index": <1-indexed int matching the preference list above>,                         
      "verbatim_text": "<exact substring from the utterance that expresses this preference>"
    }                                         
  ]                                       
                                                                                   
  If no preferences are expressed, return: []                                                         
  Return ONLY the JSON list, no other text.                                                           
  \end{tcblisting}  

%% file: sections/a4_generation_examples.tex
\section{Qualitative Generation Examples}
\label{sec:appendix:generation_examples}

This appendix shows representative next-utterance examples from each evaluation
regime. Each block gives the rendered prompt context, followed by the aligned
human continuation and selected model continuations. User identifiers in public
forum examples are anonymized for presentation.

\subsection{Reddit}
\label{sec:appendix:examples_reddit}

\begin{tcblisting}{enhanced,breakable,colback=sub,colframe=main,boxrule=0.6pt,arc=2pt,title={Reddit qualitative example},listing only,listing options={basicstyle=\ttfamily\footnotesize,breaklines=true,columns=fullflexible,keepspaces=true}}
System:
You are simulating a Reddit user engaging with conversations with different users. You are to simulate the user u/user_a. You will be given past conversations that this user has engaged with, and based on that, continue the conversation naturally as the user. Only answer with the next user utterance and no other text.

User:
Here is the conversation context, consisting of Reddit thread(s):

###THREADS START

[Title]: Why the hate

[Original Poster (OP) u/user_a]: Why the hate with pyro mains,what did we ever do?

[u/user_b]: Because pyro is borderline OP and it takes the least amount of skill to perform very well with it. It's DPS is ridiculous, flames catch you 360 degrees around the pyro, no matter where he's aiming, the reach is insane, then there's the afterburn damage on top of everything. Oh, and pyros can fly now. It takes a decent player to kill a newbie pyro. People generally don't like that. So, yeah, there you go.

[Original Poster (OP) u/user_a]: well i fly,and i died many times because i suck

[u/user_b]: Everyone dies many times. It's a game. And with pyro simple w+m1 works annoyingly well. Just don't charge a heavy from 5km away while shooting flares at him.

[Original Poster (OP) u/user_a]: i dont use flares,i use the jetpack and tey to get behind them with bacnburner

[u/user_b]: All you have to do is be a little bit sneaky. Backburner melts everything. TF2 is meant to be a fun game. If you get frustrated too much, just go with other class :]

[Original Poster (OP) u/user_a]:

###THREADS END

Generate a comment that is in line with the final thread conversation for user u/user_a. Continue the conversation naturally as the user. Only answer with the comment and no other text.

Human continuation:
like scout where a sniper tries to mele me when i have a force a nature at his head,when i get frustated i change to scout sniper or spy

Mistral-Small-24B-Base-2501:
Ty,maybe im just disappointed that im not better

User Agent / Qwen3-14B selector / Mistral-Small-24B-Base-2501 speaker:
"well i started from pyro and i just love spamming buttons,I have the rocket launcher secondary and i am more like a soldier pyro"
\end{tcblisting}

\subsection{WildChat-1M}
\label{sec:appendix:examples_wildchat}

\begin{tcblisting}{enhanced,breakable,colback=sub,colframe=main,boxrule=0.6pt,arc=2pt,title={WildChat-1M qualitative example},listing only,listing options={basicstyle=\ttfamily\footnotesize,breaklines=true,columns=fullflexible,keepspaces=true}}
System:
Continue the conversation naturally as the user. Only answer with the next user utterance and no other text.

User:
User: Hello

Assistant: Hello! How can I assist you today?

User: Are you chatgpt 4?

Assistant: I am an AI developed by OpenAI, based on the GPT (Generative Pretrained Transformer) architecture. Whether I'm specifically referred to as "ChatGPT-4" depends on the information and training provided by OpenAI at the time of my last update. If you have any specific questions or need assistance, feel free to ask, and I'll do my best to help!

User:

Continue the conversation naturally as the user. Only answer with the next user utterance and no other text.

Human continuation:
What version are you

Mistral-Small-24B-Base-2501:
How can we make you the me?

Mistral-Small-24B-Instruct-2501:
What can you do?
\end{tcblisting}

\subsection{LMSYS-Chat-1M}
\label{sec:appendix:examples_lmsys}

\begin{tcblisting}{enhanced,breakable,colback=sub,colframe=main,boxrule=0.6pt,arc=2pt,title={LMSYS-Chat-1M qualitative example},listing only,listing options={basicstyle=\ttfamily\footnotesize,breaklines=true,columns=fullflexible,keepspaces=true}}
System:
Continue the conversation naturally as the user. Only answer with the next user utterance and no other text.

User:
User: Hello, Who are you?

Assistant: Hello! I'm a language model called Assistant, and I was trained by researchers from Large Model Systems Organization (LMSYS).

User:

Continue the conversation naturally as the user. Only answer with the next user utterance and no other text.

Human continuation:
do you know llama.cpp?

Mistral-Small-24B-Base-2501:
What can you teach me?

Mistral-Small-24B-Instruct-2501:
Nice to meet you, Assistant. What can you do?
\end{tcblisting}

\subsection{DialOp}
\label{sec:appendix:examples_DialOp}

\begin{tcblisting}{enhanced,breakable,colback=sub,colframe=main,boxrule=0.6pt,arc=2pt,title={DialOp qualitative example},listing only,listing options={basicstyle=\ttfamily\footnotesize,breaklines=true,columns=fullflexible,keepspaces=true}}
System:
You are simulating a user engaging with conversations with a different user on planning a travel itinerary. You are to simulate the user marked as 'User'. You will be given past conversations that this user has engaged with, and based on that, you must generate a comment that is like the given user.

User:
User goals:
- family trip, best to go to places that are good for kids
- check out live music!
- keep budget below $90
- don't like: spanish, american, like: ethiopian, korean
- generally like eating vegan options
- definitely want to go to Liberty Memorial
- renting a car so preferably places with parking, but street parking also ok
- go to at least one park
- minimize travel distance

Here is the conversation context so far:

###CONVERSATION START

User: Hello

Assistant: Hello. Where do you think we should go?

User: Perhaps a place with live music could be nice.

Assistant: How about the Kozy Kar.

User: Would that be a good place for my family? I am traveling with them.

Assistant: It's good for groups so it would probably be okay.

User:

###CONVERSATION END

Generate an utterance that is in line with the final thread conversation for the User. Only answer with the utterance and no other text.

Human continuation:
Okay, that sounds good for one of the places we can visit. I only have time for two other places and I want to go to Liberty Memorial.

Mistral-Small-24B-Base-2501:
Is there a place to park?

Mistral-Small-24B-Instruct-2501:
Do you know if there is a good option for vegetarian or vegan meals as well?
\end{tcblisting}

\subsection{MultiWOZ}
\label{sec:appendix:examples_multiwoz}

\begin{tcblisting}{enhanced,breakable,colback=sub,colframe=main,boxrule=0.6pt,arc=2pt,title={MultiWOZ qualitative example},listing only,listing options={basicstyle=\ttfamily\footnotesize,breaklines=true,columns=fullflexible,keepspaces=true}}
System:
You are a simulating a user engaging in a task-oriented dialogue. Continue the conversation as the USER. Only answer with the next USER utterance and no other text.

User:
Dialogue context:
USER GOAL: You are looking for information in Cambridge. You are looking for a restaurant. The restaurant should be in the centre and should be in the expensive price range. The restaurant should serve asian oriental food. Once you find the restaurant you want to book a table for 8 people at 11:45 on sunday. Make sure you get the reference number. You are also looking for places to go in town. The attraction should be in the type of college. Make sure you get phone number, postcode, and area

--------------------

USER: Hi, yes, I'm trying to find a restaurant to go to.

SYSTEM: In what area of town would you like to dine?

USER: An expensive restraunt in town centre that serves asian oriental food

SYSTEM: Great, there's a restaurant called kymmoy that fits your criteria. Would you like their information?

USER: Can you book a table for me?

SYSTEM: Yes, what is the date and time that you will need the booking and how many people?

USER: book a table for 8 at 15:45 on sunday

SYSTEM: That time is all booked up would you like to try a different time?

USER: Yes please. Can we try 11:45?

SYSTEM: Booking was successful. The table will be reserved for 15 minutes. Reference number is : 65EX4JA4.

USER:

Continue the conversation as the USER. Only answer with the next USER utterance and no other text.

Human continuation:
Can you suggest some college's in town? I need a phone number, postcode and the area too please

Mistral-Small-24B-Base-2501:
Awesome! Well done. Can you find for me a college attraction in cambridge.

Mistral-Small-24B-Instruct-2501:
Do you think Kymmoy can accommodate large groups? I want to have the best experience for my friends and I. Also, What colleges are nearby?
\end{tcblisting}

%% file: sections/a4_metrics.tex
\section{Metric Details}
\label{sec:appendix:metric_details}

The paper reports three families of measurements. First, perplexity compares how
surprising human and generated user utterances are under the same autoregressive
model. Second, diversity metrics measure how much the multiple generated
continuations for a fixed dialogue context differ from one another. Third,
corpus-level and pragmatic metrics compare model behavior to the human corpus at
larger granularity. Unless otherwise stated, diversity metrics are computed
within each dialogue context and then averaged across contexts; context-level
means are not weighted by the number of generation pairs in that context.

\subsection{Perplexity Diagnostics}
\label{sec:appendix:ppl_metric}

Perplexity is the exponentiated average predictive uncertainty of a language
model and is a standard diagnostic for language modeling
\citep{jelinek1977perplexity}. We compute it over target tokens under an
autoregressive language model. For a target sequence $y_{1:T}$ and context $c$,
the average negative log-likelihood is
\[
  \mathrm{NLL}(y\mid c)
  =
  -\frac{1}{T}\sum_{t=1}^{T}\log p(y_t \mid c,y_{<t}),
\]
where the log is natural. The reported per-token perplexity is
\[
  \mathrm{PPL}_{\mathrm{token}}(y\mid c)
  =
  \exp\left(\frac{\sum_i \mathrm{NLL}_{i,\mathrm{total}}}
                 {\sum_i T_i}\right),
\]
with totals accumulated over all evaluated utterances. Thus average NLL is
measured in nats/token and perplexity is its exponential.

We evaluate two quantities. \emph{Human} perplexity scores the aligned human
continuation under the model. \emph{Generated} perplexity scores model-generated
continuations under the same model. For generated text, the comparison plots use
a random generation from each source context so that the generated side is a
sample from the model's distribution rather than a best-of-$N$ selection. A low
Generated perplexity together with a high Human perplexity indicates that the
model is internally self-consistent but poorly aligned to the empirical user-text
distribution.

\subsection{Lexical and Pairwise Diversity}
\label{sec:appendix:diversity_metrics}

For each dialogue context, diversity metrics are computed over the set of
generated continuations after empty outputs are removed and duplicate strings are
deduplicated. Exact duplicate multiplicity is therefore ignored in the default
diversity run; if all continuations in a context are empty or a metric is
undefined for the remaining set, that context contributes a missing value rather
than being forced to zero, except where noted below. Reported values are means
over valid contexts.

\textbf{Distinct-$n$.}
Distinct-$n$ was introduced as a response-diversity metric for neural
conversation models \citep{li2016diversitypromoting}. It is the number of
unique $n$-grams divided by the total number of $n$-grams across generated
continuations:
\[
  \mathrm{Distinct}\text{-}n
  =
  \frac{\left|\bigcup_j \mathrm{ngrams}_n(\hat{x}^{(j)})\right|}
       {\sum_j \left|\mathrm{ngrams}_n(\hat{x}^{(j)})\right|}.
\]
We report Distinct-1 and Distinct-2. Tokenization follows the evaluation
implementation's whitespace tokenization. If no $n$-grams are available after
preprocessing, the context-level score is $0$. Higher Distinct-1 or Distinct-2
means greater lexical variety, with Distinct-2 more sensitive to repeated
phrases.

\textbf{Self-BLEU.}
Self-BLEU reuses the BLEU precision-and-brevity framework
\citep{papineni2002bleu} as an intra-sample diversity metric for text
generation, following Texygen \citep{zhu2018texygen}. Each continuation is
treated as a prediction and all other continuations for that context are treated
as references. The context-level Self-BLEU score is the mean of these
continuation-level BLEU scores. We report Self-BLEU as a self-similarity metric:
lower values indicate greater within-context lexical variety, while high values
mean that the sampled continuations share much of the same surface form.

\textbf{Self-ROUGE-L.}
Self-ROUGE-L uses ROUGE-L \citep{lin2004rouge} as another pairwise
self-similarity diagnostic. For a context with $m$ deduplicated continuations,
the implementation computes ROUGE-L F-measure over generation pairs
$(\hat{x}^{(i)},\hat{x}^{(j)})$ and averages the pair scores. ROUGE-L is based on
longest common subsequence overlap, so it captures repeated subsequences even
when exact $n$-gram precision is not high. We report it as
\emph{Self-ROUGE-L}; lower values indicate more diverse generations.

\textbf{Self-BERTScore F1.}
Self-BERTScore measures semantic self-similarity rather than surface overlap
\citep{zhang2020bertscore}. The implementation scores unordered generation
pairs within a context with \texttt{microsoft/deberta-xlarge-mnli}, using no IDF
weighting. The BERTScore package's English baseline rescaling is enabled in the
raw evaluation CSVs, and the analysis report restores those values to the
ordinary unrescaled BERTScore scale before producing tables and plots. We report
the restored F1 value, \emph{Self-BERT-F1}, as the main semantic similarity
summary. Lower Self-BERT-F1 means the continuations are semantically farther
apart; values near $1$ indicate near-identical semantic content.

\subsection{Vendi Score}
\label{sec:appendix:vendi_metric}

Vendi Score measures the effective number of distinct samples represented by a
set of generations \citep{friedman2023vendi}. For a set of $m$ generated
continuations $S=\{\hat{x}^{(1)},\ldots,\hat{x}^{(m)}\}$, each continuation is
embedded with the sentence-embedding encoder used for the diversity sweep
(\texttt{sentence-transformers/all-MiniLM-L6-v2} unless otherwise specified).
Embedding vectors are $\ell_2$-normalized, and the cosine Gram matrix
$K \in \mathbb{R}^{m\times m}$ is symmetrized before eigendecomposition. Let
$\lambda_1,\ldots,\lambda_m$ be the eigenvalues of $K$, normalized so that they
sum to one. The Vendi Score is
\[
  \mathrm{Vendi}(S)
  =
  \exp\left(-\sum_{r=1}^{m}\lambda_r\log \lambda_r\right).
\]
The score can be interpreted as the effective number of semantically distinct
continuations in the sample set. A value close to one indicates that the
generations collapse to a single semantic mode, while larger values indicate
broader coverage of distinct continuations. Because Vendi is kernel-based, its
absolute value depends on the embedding model and similarity function; we
therefore compare systems under the same encoder and compute the score per
context before aggregating across contexts.

\subsection{MAUVE}
\label{sec:appendix:mauve_metric}

MAUVE \citep{pillutla2021mauve} measures corpus-level similarity between the
distribution of human continuations and the distribution of model continuations.
Unlike the self-diversity metrics above, MAUVE is not computed independently
inside each context. It compares two sets of response-only embeddings:
\[
  P = \{e(x_i^{H})\}_{i=1}^{N}, \qquad
  Q = \{e(\hat{x}_i)\}_{i=1}^{N},
\]
where $x_i^{H}$ is the aligned human continuation for context $i$ and
$\hat{x}_i$ is one sampled model continuation for the same context. Context text,
source identifiers, and human/model labels are used only for alignment and are
not embedded.

For each MAUVE resample, one generated continuation is sampled uniformly from
the available generations for every source context, producing $N$ human
embeddings and $N$ model embeddings.
We then call
\texttt{mauve.compute\_mauve} with precomputed feature matrices
\texttt{p\_features} and \texttt{q\_features}. 
The MAUVE package internally
$\ell_2$-normalizes rows, fits PCA on the combined human/model features, clusters
the projected points into histogram bins, and computes a divergence frontier
between the two empirical histograms. The reported \emph{MAUVE Human-Model}
score is the mean of the resampled MAUVE values. Higher MAUVE indicates that the
model and human continuation distributions are closer. We also record the
frontier integral and resampling spread for diagnostics, but the paper reports
MAUVE itself.

The default embedding backend for the final runs is
\texttt{gemini-embedding-2}; local smoke runs can use
\texttt{sentence-transformers/all-MiniLM-L6-v2}. The number of histogram buckets
is set consistently within a comparison run. With the automatic setting, the
package uses approximately $\max(2,\mathrm{round}(N/10))$ buckets, where $N$ is
the number of aligned contexts in the resample.

\subsection{Dialogue-Act Analysis}
\label{sec:appendix:dialogue_act_metric}

The dialogue-act analysis evaluates whether simulators preserve pragmatic
structure over task-oriented dialogue, not merely surface lexical diversity.
Dialogue acts have long been used to represent the communicative function of
conversation turns and the structure of act sequences
\citep{stolcke2000dialogue}. Recent work also uses dialogue-act structure as a
lens for analyzing LLM conversational behavior \citep{maitra2025dialogue,
shaikh2024grounding}. We focus this analysis on MultiWOZ 2.1
\citep{budzianowski2019multiwoz}, where task-oriented turns have structured
dialogue-act annotations, and use ConvLab BERTNLU
\citep{henderson2020convlab} to label generated user utterances.

BERTNLU is run with MultiWOZ context enabled: the generated utterance is labeled
using up to the previous three utterances as classifier context. If a generated
utterance yields no acts and contains multiple sentences, sentence-level
fallback labeling is used and the recovered acts are merged. The human side uses
the aligned reference turn annotations. Thus, each evaluation row compares a
human target user turn and a model replacement for the same dialogue state.

Each labeled turn is converted to a canonical intent token. We lower-case the
dialogue-act intents, sort the unique intent labels, and discard slot values for
the sequence-level analysis. For a user turn, the token has the form
\texttt{U:(intent-set)}; for a system turn, the token has the form
\texttt{S:(intent-set)}. Empty intent sets are represented by
\texttt{U:(none)} or \texttt{S:(none)}. This representation retains speaker
identity and turn-level communicative function while avoiding sparsity from
slot-value strings.

\textbf{Labeling and Sequence Construction}
Each model sequence replaces only the target human user act with the model-predicted act while preserving the reference context. For consistency, we also relabel the original human user turns over which the BERTNLU labeler achieves 88.6\% exact match accuracy.

The canonical intent inventory is \{\texttt{book}, \texttt{bye}, \texttt{greet}, \texttt{inform}, \texttt{nobook}, \texttt{nooffer}, \texttt{offerbook}, \texttt{offerbooked}, \texttt{recommend}, \texttt{reqmore}, \texttt{request}, \texttt{select}, \texttt{thank}, \texttt{welcome}\}; empty predictions are mapped to \texttt{none}. For each turn, we lower-case the detected intents, sort the unique labels, discard domain/slot/value fields for this sequence analysis, and prefix the resulting set with the speaker. Thus a user turn may become \texttt{U:(inform,request)}, while an empty system act becomes \texttt{S:(none)}. 

\textbf{Metric Definitions}
For the reasoning behind the selection of $N=1,\ldots,3$, see Appendix~\ref{sec:appendix:human_intra}. Terminal $N$-grams measure the act sequence ending at the target user turn; included $N$-grams include every local window containing that target turn. For each $N$, we construct human counts $c^H_N(g)$ and model counts $c^M_N(g)$ over the union vocabulary $\mathcal{V}_N=\mathcal{V}^H_N\cup\mathcal{V}^M_N$ of observed act $N$-grams, normalize them into empirical distributions $P_N$ and $Q_N$, and assign zero probability to patterns absent from one side. Jensen-Shannon divergence \citep{lin1991divergence} is computed between $P_N$ and $Q_N$ on this shared support. We also report top-$k$ vocabulary overlap,
$
  \mathrm{Overlap}_{N,k}
  =
  \left|\mathrm{TopK}(P_N,k)\cap \mathrm{TopK}(Q_N,k)\right| / k,
$
where $\mathrm{TopK}(P_N,k)$ is the set of the $k$ most frequent human dialogue-act $N$-grams and $\mathrm{TopK}(Q_N,k)$ is the analogous model set. Although small vocabulary sizes ($|V| < k$) naturally limit N-gram overlap, the utility of the metric remains intact, as we prioritize inter-model comparisons over absolute values across N-grams.

\textbf{Terminal $N$-grams.}
For each target user turn at index $t$, the human terminal $N$-gram is the
dialogue-act sequence ending at the human turn:
\[
  g^{H}_{t,N} = (a_{t-N+1},\ldots,a_t).
\]
The model terminal $N$-gram is formed by replacing only the final human user act
with the model-predicted user act, preserving the preceding reference context:
\[
  g^{M}_{t,N} = (a_{t-N+1},\ldots,\hat{a}_t).
\]
We compute these distributions for $N=1,\ldots,5$.

\textbf{Included $N$-grams.}
Included $N$-grams consider every length-$N$ window that contains the replaced
target turn. For the model distribution, the target token in each such window is
replaced with the model-predicted user-act token while the rest of the window is
kept from the reference dialogue. This captures how a generated act changes
local pragmatic patterns not only as the last act in a prefix, but also as part
of surrounding multi-turn context.

\textbf{Jensen-Shannon divergence.}
For a given $N$, let $P_N$ be the empirical distribution of human dialogue-act
$N$-grams and $Q_N$ be the corresponding model distribution over the union of
observed human and model patterns. We report Jensen-Shannon divergence
\citep{lin1991divergence},
\[
  \mathrm{JSD}(P_N,Q_N)
  =
  \sqrt{\frac{1}{2}D_{\mathrm{KL}}(P_N\Vert R_N)
       +\frac{1}{2}D_{\mathrm{KL}}(Q_N\Vert R_N)},
  \qquad
  R_N=\frac{1}{2}(P_N+Q_N),
\]
using the square-rooted distance returned by the SciPy Jensen-Shannon
implementation. Larger values indicate greater divergence from human pragmatic
structure.

\textbf{Top-$k$ vocabulary overlap.}
For each $N$, we compute the overlap between the $k$ most frequent human
dialogue-act $N$-grams and the $k$ most frequent model dialogue-act $N$-grams:
\[
  \mathrm{Overlap}_{N,k}
  =
  \frac{\left|\mathrm{TopK}(P_N,k)\cap \mathrm{TopK}(Q_N,k)\right|}{k}.
\]
We compute this diagnostic for $k\in\{10,50,100\}$ and report $k=50$ in the
main appendix tables. This metric asks whether a simulator recovers the common
pragmatic patterns of the human corpus, even when exact frequencies differ. We
also track model-novel pattern mass, the fraction of model $N$-gram probability
assigned to patterns never observed in the human reference distribution.

\textbf{BPE-style pattern inventory.}
As an additional sequence-level diagnostic, we train a byte-pair-encoding-style
merge inventory on human dialogue-act sequences. Starting from the canonical
speaker-marked act tokens, the most frequent adjacent pair is merged into a
macro-pattern token; this process is repeated for a fixed merge budget. The
learned human merge rules are then applied unchanged to model sequences.
Divergence, entropy ratio, and coverage ratio of the resulting macro-pattern
distributions quantify whether the model reproduces recurring multi-turn
pragmatic motifs rather than only matching isolated turn intents.

%% file: sections/a5_full_results.tex
\section{Full Result Tables}
\label{sec:appendix:full_results}

This appendix gives the complete model-by-model result tables for the reported
utterance-level metrics. The tables use the same metric definitions as
Appendix~\ref{sec:appendix:metric_details}: PPL-H scores aligned human
continuations, PPL-G$_{\mathrm{rand}}$ scores one randomly sampled generated
continuation per context, and the remaining columns report within-context
diversity or corpus-level distributional similarity. Missing cells indicate that
the metric was not available for that system family.

\subsection{Reddit}
\label{sec:appendix:full_results_reddit}

Table~\ref{tab:appendix_full_reddit} reports the full Reddit continuation
metrics for each evaluated base, instruct, reasoning, and tandem system.
\input{tables/appendix/tab_full_reddit}

\subsection{WildChat-1M}
\label{sec:appendix:full_results_wildchat}

Table~\ref{tab:appendix_full_wildchat} reports the corresponding results on
WildChat-1M human-LLM chat continuations.
\input{tables/appendix/tab_full_wildchat}

\subsection{LMSYS-Chat-1M}
\label{sec:appendix:full_results_lmsys}

Table~\ref{tab:appendix_full_lmsys} gives the full LMSYS-Chat-1M continuation
results for the same metric family.
\input{tables/appendix/tab_full_lmsys}

\subsection{MultiWOZ Continuation Metrics}

\label{sec:appendix:full_results_multiwoz_continuation}

Table~\ref{tab:appendix_full_multiwoz} reports the MultiWOZ continuation
metrics before the dialogue-act-specific analysis below.
\input{tables/appendix/tab_full_multiwoz}

\subsection{DialOp}
\label{sec:appendix:full_results_DialOp}

Table~\ref{tab:appendix_full_DialOp} gives the full DialOp continuation
results, using the same point-estimate columns as the other datasets.
\input{tables/appendix/tab_full_DialOp}

\subsection{MultiWOZ Dialogue-Act Results}
\label{sec:appendix:full_results_multiwoz}

\input{figures/appendix/indiv_da_dist}
\input{tables/appendix/tab_full_multiwoz_da}
\input{figures/appendix/da_heatmaps}

\subsubsection{Human Intrasection JSD}
\label{sec:appendix:human_intra}
\input{figures/appendix/human_bootstrapping}

%% file: tables/appendix/tab_full_reddit.tex
\begin{table*}[t]
\centering
\caption{Full Reddit point-estimate results. PPL-H scores aligned human
continuations; PPL-G$_{\mathrm{rand}}$ scores one randomly sampled generated
continuation per context.}
\small
\setlength{\tabcolsep}{3pt}
\resizebox{\textwidth}{!}{%
\begin{tabular}{lrrrrrrrrr}
\toprule
System & PPL-H & PPL-G$_{\mathrm{rand}}$ & Dist-1 & Dist-2 & Self-BLEU & Self-ROUGE-L & Self-BERT-F1 & Vendi & MAUVE \\
\midrule
Llama-3.1-8B & 13.2424 & 19.2998 & 0.6743 & 0.9602 & 0.0122 & 0.0776 & 0.4796 & 6.3915 & 0.9650 \\
Llama-3.1-8B-Instruct & 23.1513 & 6.9341 & 0.5984 & 0.9339 & 0.0545 & 0.1304 & 0.5337 & 4.7601 & 0.6946 \\
Mistral-24B-Base & 10.7243 & 17.2362 & 0.6562 & 0.9581 & 0.0152 & 0.0845 & 0.4884 & 6.2010 & 0.9658 \\
Mistral-24B-Instruct & 13.4477 & 10.4384 & 0.6147 & 0.9263 & 0.0790 & 0.1473 & 0.5560 & 5.0822 & 0.8999 \\
Qwen2.5-14B & 12.8851 & 16.4860 & 0.6770 & 0.9593 & 0.0109 & 0.0799 & 0.4831 & 6.4020 & 0.9666 \\
Qwen2.5-14B-Instruct & 32.1536 & 2.7724 & 0.5726 & 0.8520 & 0.2017 & 0.2176 & 0.6285 & 4.1150 & 0.7728 \\
Qwen2.5-7B & 14.6720 & 17.8660 & 0.6769 & 0.9598 & 0.0109 & 0.0757 & 0.4797 & 6.4709 & 0.9564 \\
Qwen2.5-7B-Instruct & 36.8406 & 2.8579 & 0.6302 & 0.8807 & 0.1697 & 0.1969 & 0.6126 & 4.4223 & 0.4143 \\
Qwen3-14B & 86.0188 & 3.2910 & 0.4960 & 0.7797 & 0.3137 & 0.2697 & 0.6519 & 3.3339 & 0.7766 \\
Qwen3-14B-Base & 13.2737 & 21.1793 & 0.6349 & 0.9503 & 0.0123 & 0.0759 & 0.4770 & 6.3468 & 0.9554 \\
Qwen3-8B & 129.8805 & 3.7846 & 0.5041 & 0.7907 & 0.2953 & 0.2620 & 0.6448 & 3.4566 & 0.7524 \\
Qwen3-8B-Base & 14.4802 & 23.2658 & 0.6747 & 0.9620 & 0.0088 & 0.0748 & 0.4806 & 6.4705 & 0.9573 \\
Tandem / Qwen3-14B / Mistral-24B-Base & -- & -- & 0.6047 & 0.9433 & 0.0244 & 0.1065 & 0.5132 & 5.4427 & 0.9343 \\
Tandem / Qwen3-14B / Qwen3-14B-Base & -- & -- & 0.5939 & 0.9394 & 0.0339 & 0.1005 & 0.5044 & 5.5744 & 0.9244 \\
Tandem / Qwen3-14B / Qwen3-14B & -- & -- & 0.4796 & 0.7698 & 0.2789 & 0.2763 & 0.6573 & 3.1051 & 0.7272 \\
\bottomrule
\end{tabular}
}
\label{tab:appendix_full_reddit}
\end{table*}

%% file: tables/appendix/tab_full_wildchat.tex
\begin{table*}[t]
\centering
\caption{Full WildChat-1M point-estimate results. PPL-H scores aligned human
continuations; PPL-G$_{\mathrm{rand}}$ scores one randomly sampled generated
continuation per context.}
\small
\setlength{\tabcolsep}{3pt}
\resizebox{\textwidth}{!}{%
\begin{tabular}{lrrrrrrrrr}
\toprule
System & PPL-H & PPL-G$_{\mathrm{rand}}$ & Dist-1 & Dist-2 & Self-BLEU & Self-ROUGE-L & Self-BERT-F1 & Vendi & MAUVE \\
\midrule
Llama-3.1-8B & 4.3322 & 8.1718 & 0.6483 & 0.8977 & 0.0785 & 0.1122 & 0.5141 & 6.1210 & 0.8608 \\
Llama-3.1-8B-Instruct & 5.2963 & 9.5169 & 0.6068 & 0.9044 & 0.0865 & 0.1510 & 0.5840 & 4.7968 & 0.7813 \\
Mistral-24B-Base & 3.5866 & 8.9617 & 0.6321 & 0.8875 & 0.0902 & 0.1170 & 0.5166 & 5.9821 & 0.9659 \\
Mistral-24B-Instruct & 4.0625 & 7.0132 & 0.6758 & 0.9275 & 0.0598 & 0.1315 & 0.5812 & 5.5360 & 0.6233 \\
Qwen2.5-14B & 3.8005 & 7.4436 & 0.6463 & 0.8871 & 0.0922 & 0.1114 & 0.5102 & 6.0735 & 0.9691 \\
Qwen2.5-14B-Instruct & 5.8571 & 2.4996 & 0.5625 & 0.7978 & 0.2501 & 0.2386 & 0.6641 & 4.0393 & 0.5007 \\
Qwen2.5-7B & 4.0866 & 6.8526 & 0.6406 & 0.8860 & 0.0934 & 0.1135 & 0.5118 & 5.9729 & 0.9637 \\
Qwen2.5-7B-Instruct & 6.2130 & 2.9001 & 0.5402 & 0.7549 & 0.3020 & 0.2791 & 0.6924 & 3.9961 & 0.4634 \\
Qwen3-14B & 6.3684 & 1.6479 & 0.4825 & 0.7123 & 0.3968 & 0.3021 & 0.6781 & 3.2709 & 0.8739 \\
Qwen3-14B-Base & 3.7557 & 5.4252 & 0.6338 & 0.8815 & 0.0934 & 0.1116 & 0.5094 & 6.0090 & 0.9682 \\
Qwen3-8B & 7.0478 & 1.7676 & 0.5031 & 0.7289 & 0.3682 & 0.2869 & 0.6689 & 3.3980 & 0.8489 \\
Qwen3-8B-Base & 3.9346 & 9.8134 & 0.6287 & 0.8807 & 0.0937 & 0.1101 & 0.5097 & 6.0507 & 0.9696 \\
Tandem / Qwen3-14B / Mistral-24B-Base & -- & -- & 0.5752 & 0.8527 & 0.1318 & 0.1577 & 0.5505 & 5.0599 & 0.9672 \\
Tandem / Qwen3-14B / Qwen3-14B-Base & -- & -- & 0.5741 & 0.8439 & 0.1523 & 0.1588 & 0.5491 & 4.9824 & 0.9746 \\
\bottomrule
\end{tabular}
}
\label{tab:appendix_full_wildchat}
\end{table*}

%% file: tables/appendix/tab_full_lmsys.tex
\begin{table*}[t]
\centering
\caption{Full LMSYS-Chat-1M point-estimate results. PPL-H scores aligned human
continuations; PPL-G$_{\mathrm{rand}}$ scores one randomly sampled generated
continuation per context.}
\small
\setlength{\tabcolsep}{3pt}
\resizebox{\textwidth}{!}{%
\begin{tabular}{lrrrrrrrrr}
\toprule
System & PPL-H & PPL-G$_{\mathrm{rand}}$ & Dist-1 & Dist-2 & Self-BLEU & Self-ROUGE-L & Self-BERT-F1 & Vendi & MAUVE \\
\midrule
Llama-3.1-8B & 4.0134 & 7.1166 & 0.6723 & 0.8993 & 0.0660 & 0.1117 & 0.5177 & 6.3393 & 0.8206 \\
Llama-3.1-8B-Instruct & 5.8418 & 3.5852 & 0.5955 & 0.8857 & 0.1070 & 0.1662 & 0.5911 & 4.8261 & 0.7639 \\
Mistral-24B-Base & 3.3318 & 7.3487 & 0.6714 & 0.8959 & 0.0746 & 0.1162 & 0.5185 & 6.2141 & 0.9503 \\
Mistral-24B-Instruct & 4.3267 & 4.9720 & 0.6673 & 0.9010 & 0.0918 & 0.1625 & 0.5977 & 5.3387 & 0.7026 \\
Qwen2.5-14B & 3.4565 & 6.1836 & 0.6676 & 0.8830 & 0.0837 & 0.1194 & 0.5196 & 6.1912 & 0.9499 \\
Qwen2.5-14B-Instruct & 12.5903 & 2.1503 & 0.5200 & 0.7306 & 0.3556 & 0.3049 & 0.6864 & 3.6589 & 0.6036 \\
Qwen2.5-7B & 3.6256 & 6.8400 & 0.6791 & 0.8942 & 0.0825 & 0.1156 & 0.5195 & 6.2180 & 0.9499 \\
Qwen2.5-7B-Instruct & 10.1829 & 2.1135 & 0.5207 & 0.7123 & 0.3595 & 0.3150 & 0.7053 & 3.9144 & 0.5897 \\
Qwen3-14B & 11.6769 & 2.0274 & 0.4873 & 0.6847 & 0.4276 & 0.3369 & 0.6913 & 3.1596 & 0.8327 \\
Qwen3-14B-Base & 3.3858 & 6.2173 & 0.6669 & 0.8892 & 0.0772 & 0.1043 & 0.5063 & 6.3051 & 0.9542 \\
Qwen3-8B & 13.1645 & 2.3498 & 0.5127 & 0.7134 & 0.3784 & 0.3075 & 0.6790 & 3.4498 & 0.8237 \\
Qwen3-8B-Base & 3.5010 & 7.9230 & 0.6708 & 0.8996 & 0.0719 & 0.1049 & 0.5065 & 6.3690 & 0.9516 \\
Tandem / Qwen3-14B / Mistral-24B-Base & -- & -- & 0.6217 & 0.8653 & 0.1162 & 0.1603 & 0.5574 & 5.2694 & 0.9421 \\
Tandem / Qwen3-14B / Qwen3-14B-Base & -- & -- & 0.5992 & 0.8459 & 0.1469 & 0.1606 & 0.5512 & 5.2067 & 0.9587 \\
Tandem / Qwen3-14B / Qwen3-14B & -- & -- & 0.4464 & 0.6482 & 0.4367 & 0.3714 & 0.7109 & 2.8156 & 0.8054 \\
\bottomrule
\end{tabular}
}
\label{tab:appendix_full_lmsys}
\end{table*}

%% file: tables/appendix/tab_full_multiwoz.tex
\begin{table*}[t]
\centering
\caption{Full MultiWOZ continuation point-estimate results. PPL-H scores aligned
human continuations; PPL-G$_{\mathrm{rand}}$ scores one randomly sampled
generated continuation per context.}
\small
\setlength{\tabcolsep}{3pt}
\resizebox{\textwidth}{!}{%
\begin{tabular}{lrrrrrrrrr}
\toprule
System & PPL-H & PPL-G$_{\mathrm{rand}}$ & Dist-1 & Dist-2 & Self-BLEU & Self-ROUGE-L & Self-BERT-F1 & Vendi & MAUVE \\
\midrule
Llama-3.1-8B & 5.5176 & 7.9276 & 0.7019 & 0.9066 & 0.0809 & 0.1862 & 0.5428 & 5.1383 & 0.6172 \\
Llama-3.1-8B-Instruct & 10.7274 & 2.8377 & 0.5258 & 0.8028 & 0.2477 & 0.2610 & 0.6700 & 3.6958 & 0.3526 \\
Mistral-24B-Base & 4.8951 & 5.8217 & 0.6910 & 0.8948 & 0.0956 & 0.1845 & 0.5804 & 5.0722 & 0.6743 \\
Mistral-24B-Instruct & 5.0876 & 3.6643 & 0.5731 & 0.8003 & 0.2424 & 0.2851 & 0.6932 & 3.8587 & 0.6687 \\
Qwen2.5-14B & 5.1235 & 6.3018 & 0.6654 & 0.8824 & 0.1153 & 0.2005 & 0.6367 & 4.9447 & 0.7132 \\
Qwen2.5-14B-Instruct & 20.0783 & 1.7114 & 0.4251 & 0.6084 & 0.5541 & 0.4712 & 0.6949 & 2.5738 & 0.7499 \\
Qwen2.5-7B & 5.2737 & 6.6054 & 0.6811 & 0.8959 & 0.0990 & 0.1879 & 0.5992 & 5.0275 & 0.7100 \\
Qwen2.5-7B-Instruct & 27.2847 & 1.5299 & 0.4089 & 0.5647 & 0.5995 & 0.5266 & 0.7808 & 2.4876 & 0.7171 \\
Qwen3-14B & 72.2336 & 2.7179 & 0.4119 & 0.5431 & 0.6375 & 0.5187 & 0.7802 & 2.3100 & 0.5098 \\
Qwen3-14B-Base & 5.0428 & 5.7457 & 0.6600 & 0.8800 & 0.1162 & 0.1898 & 0.5905 & 5.0127 & 0.7116 \\
Qwen3-8B & 65.4066 & 2.8170 & 0.4608 & 0.5839 & 0.5689 & 0.4950 & 0.9343 & 2.4334 & 0.5816 \\
Qwen3-8B-Base & 5.2780 & 6.3251 & 0.6829 & 0.8939 & 0.1026 & 0.1809 & 0.4689 & 5.1156 & 0.6535 \\
Tandem / Qwen3-14B / Mistral-24B-Base & -- & -- & 0.5884 & 0.8143 & 0.2177 & 0.2912 & 0.6641 & 3.9454 & 0.8664 \\
Tandem / Qwen3-14B / Qwen3-14B-Base & -- & -- & 0.5692 & 0.8062 & 0.2140 & 0.2880 & 0.6590 & 3.9370 & 0.4029 \\
Tandem / Qwen3-14B / Qwen3-14B & -- & -- & 0.3541 & 0.4872 & 0.6496 & 0.5773 & 0.8085 & 2.0580 & 0.0957 \\
\bottomrule
\end{tabular}
}
\label{tab:appendix_full_multiwoz}
\end{table*}

%% file: tables/appendix/tab_full_dialop.tex
\begin{table*}[t]
\centering
\caption{Full DialOp point-estimate results. PPL-H scores aligned human
continuations; PPL-G$_{\mathrm{rand}}$ scores one randomly sampled generated
continuation per context.}
\small
\setlength{\tabcolsep}{3pt}
\resizebox{\textwidth}{!}{%
\begin{tabular}{lrrrrrrrrr}
\toprule
System & PPL-H & PPL-G$_{\mathrm{rand}}$ & Dist-1 & Dist-2 & Self-BLEU & Self-ROUGE-L & Self-BERT-F1 & Vendi & MAUVE \\
\midrule
Llama-3.1-8B & 12.4540 & 16.5786 & 0.7327 & 0.9570 & 0.0125 & 0.0829 & 0.5614 & 6.5064 & 0.4638 \\
Llama-3.1-8B-Instruct & 23.6024 & 4.4736 & 0.5588 & 0.8957 & 0.0868 & 0.1711 & 0.6723 & 4.1654 & 0.3901 \\
Mistral-24B-Base & 10.9471 & 12.0142 & 0.7194 & 0.9522 & 0.0180 & 0.0821 & 0.5996 & 6.4174 & 0.5942 \\
Mistral-24B-Instruct & 12.3744 & 9.9723 & 0.5982 & 0.9307 & 0.0434 & 0.1367 & 0.6078 & 4.8471 & 0.4900 \\
Qwen2.5-14B & 12.3213 & 8.0706 & 0.7019 & 0.9454 & 0.0245 & 0.0911 & 0.5743 & 6.2673 & 0.5640 \\
Qwen2.5-14B-Instruct & 57.3444 & 2.4584 & 0.5177 & 0.7909 & 0.2729 & 0.2582 & 0.6922 & 3.4711 & 0.3958 \\
Qwen2.5-7B & 12.9216 & 10.9312 & 0.7010 & 0.9458 & 0.0207 & 0.0833 & 0.5352 & 6.3224 & 0.5293 \\
Qwen2.5-7B-Instruct & 73.1071 & 2.2353 & 0.5127 & 0.7884 & 0.2967 & 0.2657 & 0.7231 & 3.2720 & 0.2140 \\
Qwen3-14B & 144.0394 & 4.5477 & 0.4180 & 0.6866 & 0.4523 & 0.3576 & 0.7450 & 2.3924 & 0.0835 \\
Qwen3-14B-Base & 11.9030 & 10.3060 & 0.7005 & 0.9455 & 0.0153 & 0.0822 & 0.5151 & 6.3563 & 0.6440 \\
Qwen3-8B & 194.7983 & 4.5915 & 0.4189 & 0.7103 & 0.3920 & 0.3205 & 0.7336 & 2.4051 & 0.0904 \\
Qwen3-8B-Base & 12.8801 & 8.3997 & 0.6923 & 0.9323 & 0.0216 & 0.0782 & 0.4444 & 6.3152 & 0.5132 \\
Tandem / Qwen3-14B / Mistral-24B-Base & -- & -- & 0.6780 & 0.9362 & 0.0379 & 0.1091 & 0.5519 & 5.7922 & 0.5669 \\
\bottomrule
\end{tabular}
}
\label{tab:appendix_full_DialOp}
\end{table*}

%% file: figures/appendix/indiv_da_dist.tex
\begin{figure}[t]
\centering
\includegraphics[height=0.9\textheight, width=\linewidth, keepaspectratio]{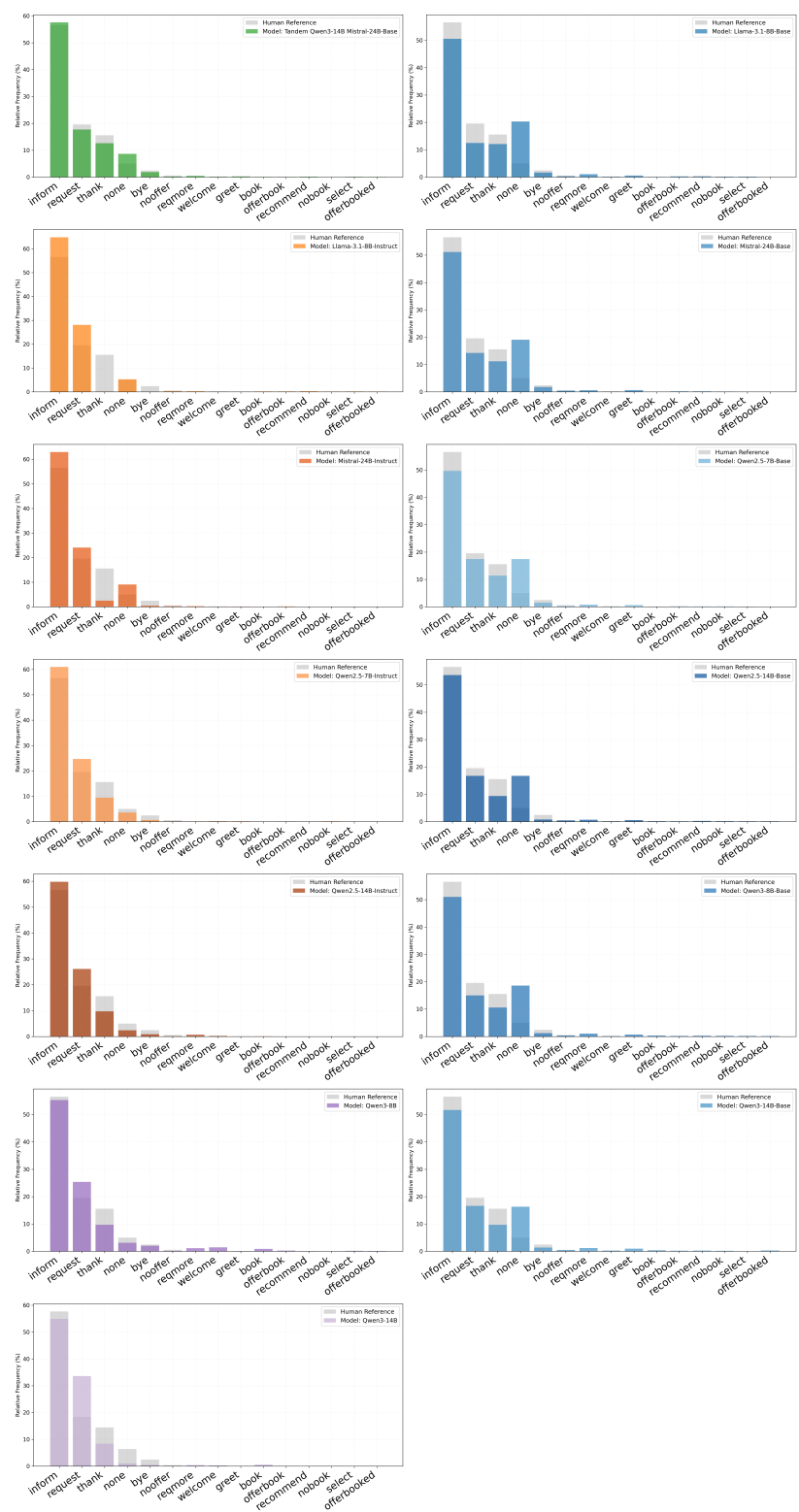}
\caption{Individual (N=1) dialogue act distributions for each model compared against the human distribution.}
\label{fig:indiv_da_dist}
\end{figure}

%% file: tables/appendix/tab_full_multiwoz_da.tex
\begin{table*}[t]
\centering
\caption{Comprehensive results for N-gram Dialogue Act sequences ($N=1$ to $3$). Results include Jensen-Shannon Divergence (JSD $\downarrow$) and Vocabulary Overlap ($\uparrow$) at multiple $K$ thresholds.}
\small
\setlength{\tabcolsep}{2.5pt}
\resizebox{\textwidth}{!}{%
\begin{tabular}{l ccc c ccc c ccc c ccc}
\toprule
& \multicolumn{3}{c}{JSD ($\downarrow$)} & & \multicolumn{3}{c}{Overlap K=10 ($\uparrow$)} & & \multicolumn{3}{c}{Overlap K=100 ($\uparrow$)} & & \multicolumn{3}{c}{Overlap K=200 ($\uparrow$)} \\
\cmidrule{2-4} \cmidrule{6-8} \cmidrule{10-12} \cmidrule{14-16}
System & $N=1$ & $2$ & $3$ & & $N=1$ & $2$ & $3$ & & $N=1$ & $2$ & $3$ & & $N=1$ & $2$ & $3$ \\
\midrule
Llama-3.1-8B-Base     & 0.198 & 0.293 & 0.346 & & 0.700 & 0.600 & 0.600 & & 0.080 & 0.700 & 0.770 & & 0.040 & 0.600 & 0.645 \\
Llama-3.1-8B-Inst     & 0.265 & 0.366 & 0.407 & & 0.400 & 0.700 & 0.700 & & 0.070 & 0.640 & 0.640 & & 0.035 & 0.495 & 0.570 \\
\midrule
Mistral-24B-Base      & 0.181 & 0.277 & 0.330 & & 0.700 & \textbf{0.800} & 0.500 & & 0.080 & 0.750 & 0.740 & & 0.040 & 0.620 & 0.660 \\
Mistral-24B-Inst      & 0.195 & 0.299 & 0.346 & & 0.700 & 0.700 & \textbf{0.800} & & 0.080 & 0.680 & 0.670 & & 0.040 & 0.575 & 0.635 \\
\midrule
Qwen2.5-7B-Base       & 0.168 & 0.271 & 0.325 & & 0.700 & \textbf{0.800} & 0.600 & & 0.080 & \textbf{0.760} & 0.780 & & 0.040 & 0.635 & 0.660 \\
Qwen2.5-7B-Inst       & 0.107 & 0.258 & 0.310 & & 0.700 & \textbf{0.800} & 0.700 & & 0.080 & 0.690 & 0.740 & & 0.040 & 0.560 & 0.665 \\
\midrule
Qwen2.5-14B-Base      & 0.175 & 0.272 & 0.325 & & 0.700 & \textbf{0.800} & 0.600 & & 0.080 & 0.680 & 0.760 & & 0.040 & 0.610 & 0.665 \\
Qwen2.5-14B-Inst      & 0.119 & 0.273 & 0.321 & & 0.700 & 0.700 & 0.600 & & 0.080 & 0.680 & 0.750 & & 0.040 & 0.585 & 0.635 \\
\midrule
Qwen3-8B-Base         & 0.190 & 0.287 & 0.338 & & 0.700 & 0.700 & 0.500 & & 0.080 & 0.730 & 0.720 & & 0.040 & 0.620 & 0.635 \\
Qwen3-8B              & 0.138 & 0.299 & 0.343 & & 0.700 & \textbf{0.800} & 0.600 & & 0.080 & 0.650 & 0.760 & & 0.040 & 0.520 & 0.595 \\
\midrule
Qwen3-14B-Base        & 0.180 & 0.282 & 0.331 & & 0.700 & 0.700 & 0.600 & & 0.080 & 0.710 & 0.770 & & 0.040 & 0.590 & \textbf{0.670} \\
Qwen3-14B             & 0.184 & 0.354 & 0.389 & & 0.700 & 0.400 & 0.300 & & 0.080 & 0.650 & 0.670 & & 0.040 & 0.545 & 0.565 \\
\midrule
Tandem (Q3/M24)       & \textbf{0.088} & \textbf{0.219} & \textbf{0.280} & & \textbf{0.800} & \textbf{0.800} & \textbf{0.800} & & 0.080 & 0.750 & \textbf{0.840} & & 0.040 & \textbf{0.650} & 0.665 \\
\bottomrule
\end{tabular}
}
\label{tab:combined_da_metrics}
\end{table*}

%% file: figures/appendix/da_heatmaps.tex
\begin{figure}[ht]
\centering
\includegraphics[width=\linewidth]{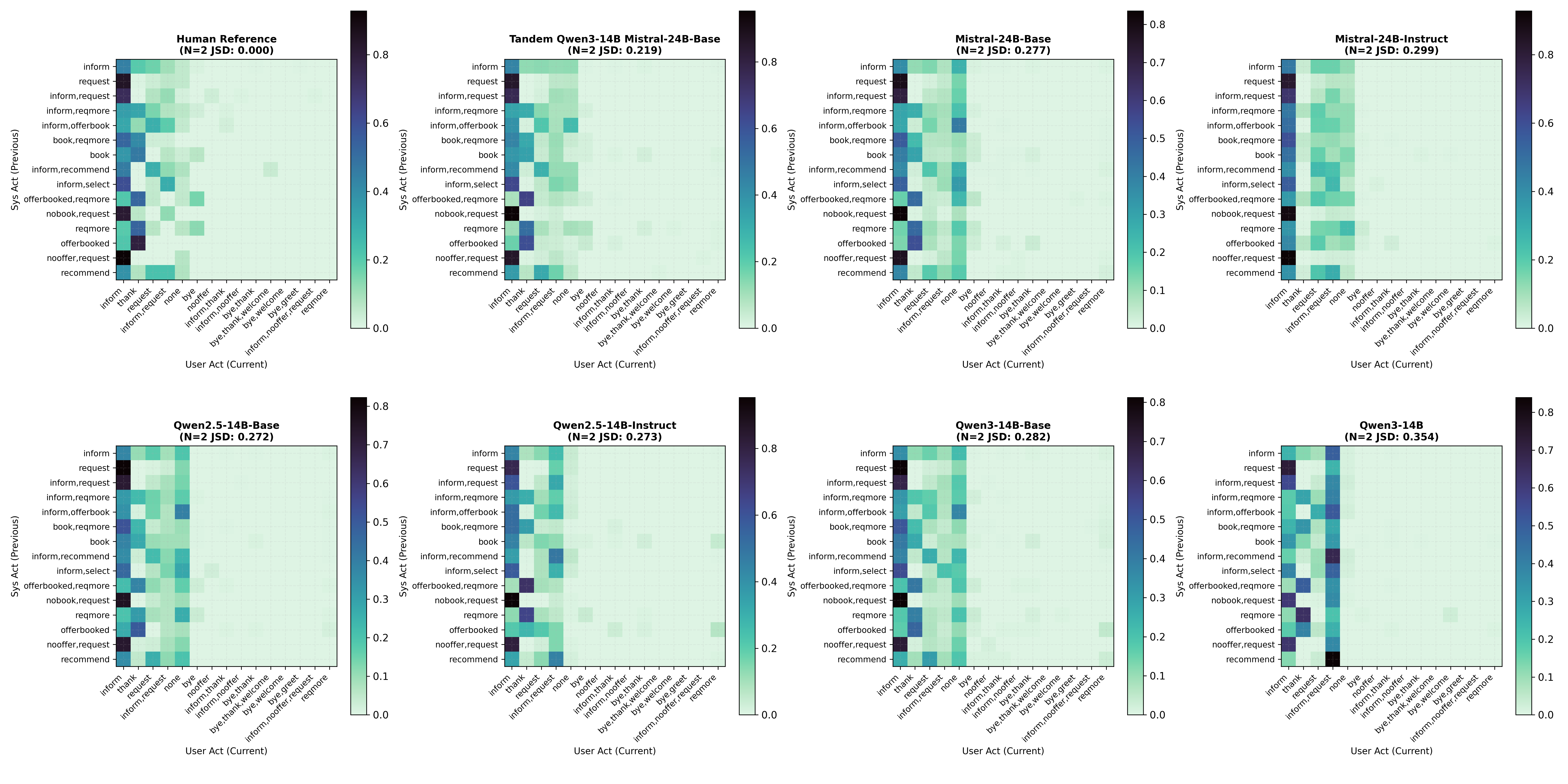}
\caption{Transition probabilities for N=2 (i.e., probability of user intent given previous assistant intent).}
\label{fig:da_heatmaps}
\end{figure}

%% file: figures/appendix/human_bootstrapping.tex
\begin{figure}[ht]
    \centering
    \begin{minipage}[b]{0.45\linewidth}
        \centering
        \includegraphics[width=\textwidth]{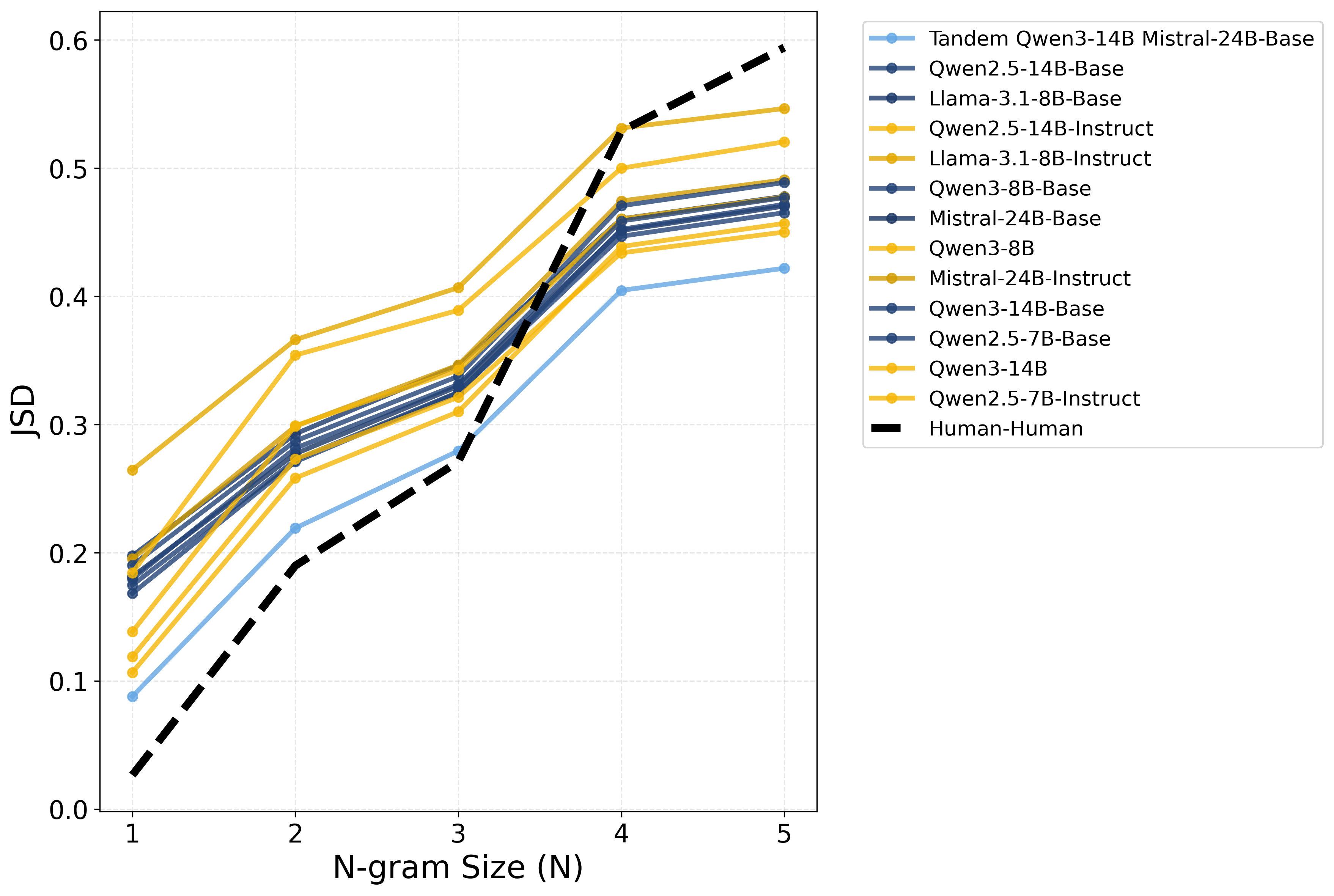}
    \end{minipage}
    \hfill 
    \begin{minipage}[b]{0.51\linewidth}
        \centering
        \includegraphics[width=\textwidth]{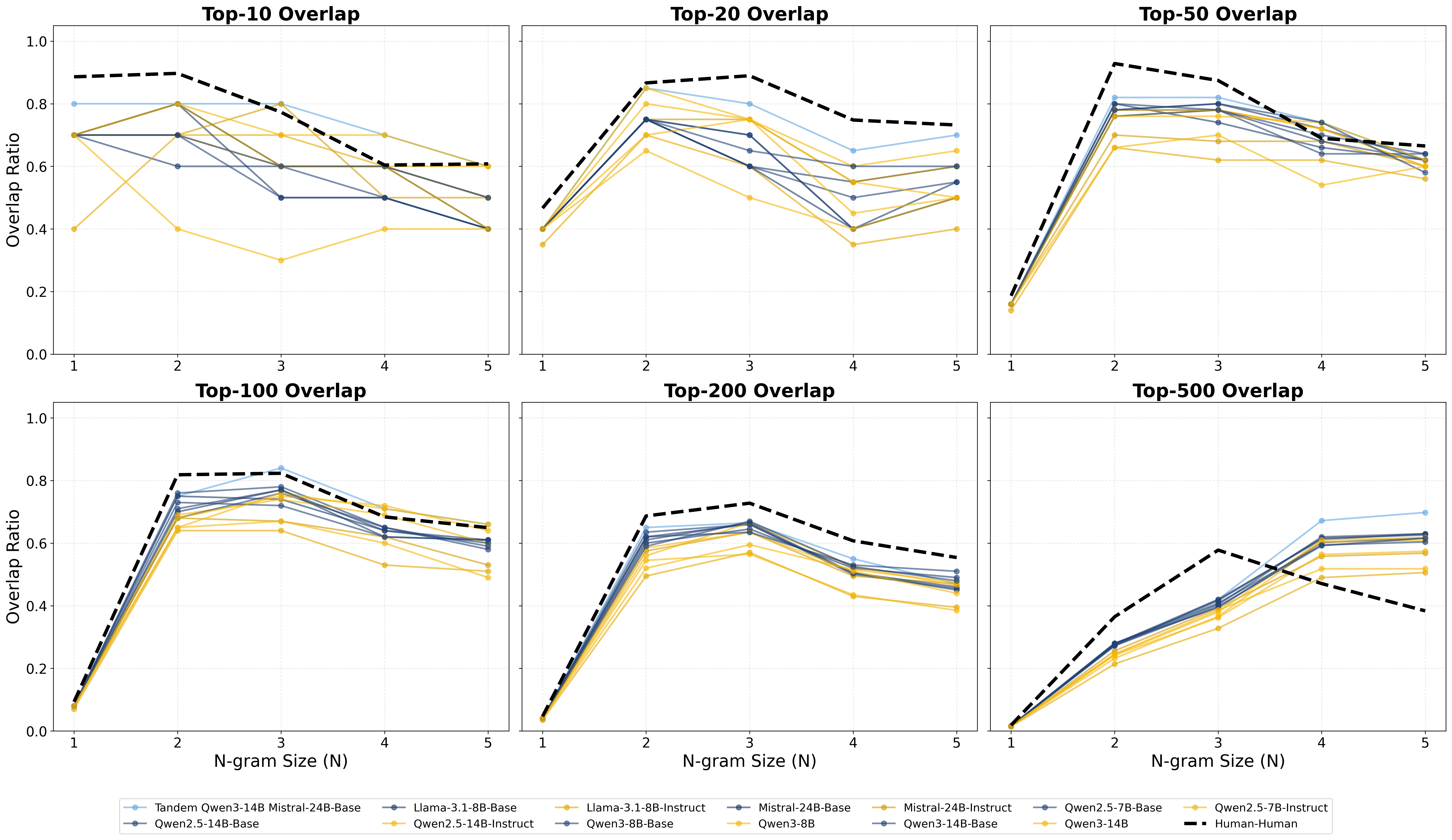}
    \end{minipage}
    
    \vspace{5pt} 
    \caption{Comparison of distributional divergence and vocabulary overlap between model outputs and human-human (H-H) bootstrapped baselines. H-H results are generated via 100$\times$ random corpus splits. We observe that for $N \leq 3$, H-H performance consistently exceeds model-human similarity, establishing $N=3$ as our threshold complexity limit.}
    \label{fig:side_by_side}
\end{figure}

%% file: higgins.bib
@inproceedings{chang2020convokit,
	title        = {{C}onvo{K}it: A Toolkit for the Analysis of Conversations},
	author       = {Chang, Jonathan P.  and Chiam, Caleb  and Fu, Liye  and Wang, Andrew  and Zhang, Justine  and Danescu-Niculescu-Mizil, Cristian},
	year         = 2020,
	booktitle    = {Proceedings of the 21th Annual Meeting of the Special Interest Group on Discourse and Dialogue},
	publisher    = {Association for Computational Linguistics},
	address      = {1st virtual meeting},
	pages        = {57--60},
	doi          = {10.18653/v1/2020.sigdial-1.8},
	url          = {https://aclanthology.org/2020.sigdial-1.8/},
	editor       = {Pietquin, Olivier  and Muresan, Smaranda  and Chen, Vivian  and Kennington, Casey  and Vandyke, David  and Dethlefs, Nina  and Inoue, Koji  and Ekstedt, Erik  and Ultes, Stefan}
}

@inproceedings{zhao2024wildchat,
	title        = {WildChat: 1M ChatGPT Interaction Logs in the Wild},
	author       = {Wenting Zhao and Xiang Ren and Jack Hessel and Claire Cardie and Yejin Choi and Yuntian Deng},
	year         = 2024,
	booktitle    = {The Twelfth International Conference on Learning Representations, {ICLR} 2024, Vienna, Austria, May 7-11, 2024},
	publisher    = {OpenReview.net},
	url          = {https://openreview.net/forum?id=Bl8u7ZRlbM},
	bibsource    = {dblp computer science bibliography, https://dblp.org}
}

@inproceedings{zheng2023lmsys,
	title        = {LMSYS-Chat-1M: {A} Large-Scale Real-World {LLM} Conversation Dataset},
	author       = {Lianmin Zheng and Wei{-}Lin Chiang and Ying Sheng and Tianle Li and Siyuan Zhuang and Zhanghao Wu and Yonghao Zhuang and Zhuohan Li and Zi Lin and Eric P. Xing and Joseph E. Gonzalez and Ion Stoica and Hao Zhang},
	year         = 2024,
	booktitle    = {The Twelfth International Conference on Learning Representations, {ICLR} 2024, Vienna, Austria, May 7-11, 2024},
	publisher    = {OpenReview.net},
	url          = {https://openreview.net/forum?id=BOfDKxfwt0},
	bibsource    = {dblp computer science bibliography, https://dblp.org}
}

@article{lin2024decision,
	title        = {Decision-Oriented Dialogue for Human-{AI} Collaboration},
	author       = {Lin, Jessy  and Tomlin, Nicholas  and Andreas, Jacob  and Eisner, Jason},
	year         = 2024,
	journal      = {Transactions of the Association for Computational Linguistics},
	publisher    = {MIT Press},
	address      = {Cambridge, MA},
	volume       = 12,
	pages        = {892--911},
	doi          = {10.1162/tacl_a_00679},
	url          = {https://aclanthology.org/2024.tacl-1.50/}
}

@inproceedings{budzianowski2019multiwoz,
	title        = {{M}ulti{WOZ} - A Large-Scale Multi-Domain {W}izard-of-{O}z Dataset for Task-Oriented Dialogue Modelling},
	author       = {Budzianowski, Pawe{\l}  and Wen, Tsung-Hsien  and Tseng, Bo-Hsiang  and Casanueva, I{\~n}igo  and Ultes, Stefan  and Ramadan, Osman  and Ga{\v{s}}i{\'c}, Milica},
	year         = 2018,
	booktitle    = {Proceedings of the 2018 Conference on Empirical Methods in Natural Language Processing},
	publisher    = {Association for Computational Linguistics},
	address      = {Brussels, Belgium},
	pages        = {5016--5026},
	doi          = {10.18653/v1/D18-1547},
	url          = {https://aclanthology.org/D18-1547/},
	editor       = {Riloff, Ellen  and Chiang, David  and Hockenmaier, Julia  and Tsujii, Jun{'}ichi}
}

@inproceedings{henderson2020convlab,
	title        = {{C}onv{L}ab-2: An Open-Source Toolkit for Building, Evaluating, and Diagnosing Dialogue Systems},
	author       = {Zhu, Qi  and Zhang, Zheng  and Fang, Yan  and Li, Xiang  and Takanobu, Ryuichi  and Li, Jinchao  and Peng, Baolin  and Gao, Jianfeng  and Zhu, Xiaoyan  and Huang, Minlie},
	year         = 2020,
	booktitle    = {Proceedings of the 58th Annual Meeting of the Association for Computational Linguistics: System Demonstrations},
	publisher    = {Association for Computational Linguistics},
	address      = {Online},
	pages        = {142--149},
	doi          = {10.18653/v1/2020.acl-demos.19},
	url          = {https://aclanthology.org/2020.acl-demos.19/},
	editor       = {Celikyilmaz, Asli  and Wen, Tsung-Hsien}
}

@inproceedings{naous2026flipping,
	title        = {Flipping the Dialogue: Training and Evaluating User Language Models},
	author       = {Naous, Tarek and Laban, Philippe and Xu, Wei and Neville, Jennifer},
	year         = 2026,
	booktitle    = {The Fourteenth International Conference on Learning Representations},
	note         = {arXiv:2510.06552}
}

@inproceedings{seshadri2026lost,
	title        = {Lost in Simulation: {LLM}-Simulated Users are Unreliable Proxies for Human Users in Agentic Evaluations},
	author       = {Seshadri, Preethi  and Cahyawijaya, Samuel  and Odumakinde, Ayomide  and Singh, Sameer  and Goldfarb-Tarrant, Seraphina},
	year         = 2026,
	booktitle    = {Proceedings of the 64th Annual Meeting of the {A}ssociation for {C}omputational {L}inguistics (Volume 1: Long Papers)},
	publisher    = {Association for Computational Linguistics},
	address      = {San Diego, California, United States},
	pages        = {47423--47439},
	doi          = {10.18653/v1/2026.acl-long.2192},
	isbn         = {979-8-89176-390-6},
	url          = {https://aclanthology.org/2026.acl-long.2192/},
	editor       = {Liakata, Maria  and Moreira, Viviane P.  and Zhang, Jiajun  and Jurgens, David}
}

@inproceedings{go2023aligning,
	title        = {Aligning Language Models with Preferences through f-divergence Minimization},
	author       = {Dongyoung Go and Tomasz Korbak and Germ{\'{a}}n Kruszewski and Jos Rozen and Nahyeon Ryu and Marc Dymetman},
	year         = 2023,
	booktitle    = {International Conference on Machine Learning, {ICML} 2023, 23-29 July 2023, Honolulu, Hawaii, {USA}},
	publisher    = {{PMLR}},
	series       = {Proceedings of Machine Learning Research},
	volume       = 202,
	pages        = {11546--11583},
	url          = {https://proceedings.mlr.press/v202/go23a.html},
	bibsource    = {dblp computer science bibliography, https://dblp.org},
	editor       = {Andreas Krause and Emma Brunskill and Kyunghyun Cho and Barbara Engelhardt and Sivan Sabato and Jonathan Scarlett}
}

@misc{wu2026humanlm,
	title        = {{HumanLM}: Simulating Users with State Alignment Beats Response Imitation},
	author       = {Wu, Shirley and Choi, Evelyn and Khatua, Arpandeep and Wang, Zhanghan and He-Yueya, Joy and Weerasooriya, Tharindu Cyril and Wei, Wei and Yang, Diyi and Leskovec, Jure and Zou, James},
	year         = 2026,
	url          = {https://arxiv.org/abs/2603.03303},
	archiveprefix = {arXiv},
	eprint       = {2603.03303},
	primaryclass = {cs.CL}
}

@inproceedings{abdulhai2025consistently,
	title        = {Consistently Simulating Human Personas with Multi-Turn Reinforcement Learning},
	author       = {Marwa Abdulhai and Ryan Cheng and Donovan Clay and Tim Althoff and Sergey Levine and Natasha Jaques},
	year         = 2025,
	booktitle    = {Advances in Neural Information Processing Systems 38: Annual Conference on Neural Information Processing Systems 2025, NeurIPS 2025, San Diego, CA, USA, December 2-7, 2025 / Mexico City, Mexico, November 30 - December 5, 2025},
	url          = {http://papers.nips.cc/paper\_files/paper/2025/hash/4c91443877f8388d8190c938ac5a4d4d-Abstract-Conference.html},
	bibsource    = {dblp computer science bibliography, https://dblp.org},
	editor       = {Danielle Belgrave and Cheng Zhang and Laura N. Montoya and Hsuan{-}Tien Lin and Razvan Pascanu and Piotr Koniusz and Marzyeh Ghassemi and Nancy Chen and Iv{\'{a}}n Vladimir Meza Ru{\'{\i}}z and Arturo Loaiza{-}Bonilla}
}

@inproceedings{kazi2024llm,
	title        = {Large language models as user-agents for evaluating task-oriented-dialogue systems},
	author       = {Kazi, Taaha and Lyu, Ruiliang and Zhou, Sizhe and Hakkani-T{\"u}r, Dilek and Tur, Gokhan},
	year         = 2024,
	booktitle    = {2024 IEEE Spoken Language Technology Workshop (SLT)},
	pages        = {913--920},
	organization = {IEEE}
}

@inproceedings{pillutla2021mauve,
	title        = {{MAUVE:} Measuring the Gap Between Neural Text and Human Text using Divergence Frontiers},
	author       = {Krishna Pillutla and Swabha Swayamdipta and Rowan Zellers and John Thickstun and Sean Welleck and Yejin Choi and Za{\"{\i}}d Harchaoui},
	year         = 2021,
	booktitle    = {Advances in Neural Information Processing Systems 34: Annual Conference on Neural Information Processing Systems 2021, NeurIPS 2021, December 6-14, 2021, virtual},
	pages        = {4816--4828},
	url          = {https://proceedings.neurips.cc/paper/2021/hash/260c2432a0eecc28ce03c10dadc078a4-Abstract.html},
	bibsource    = {dblp computer science bibliography, https://dblp.org},
	editor       = {Marc'Aurelio Ranzato and Alina Beygelzimer and Yann N. Dauphin and Percy Liang and Jennifer Wortman Vaughan}
}

@inproceedings{lin2024urial,
	title        = {The Unlocking Spell on Base LLMs: Rethinking Alignment via In-Context Learning},
	author       = {Bill Yuchen Lin and Abhilasha Ravichander and Ximing Lu and Nouha Dziri and Melanie Sclar and Khyathi Raghavi Chandu and Chandra Bhagavatula and Yejin Choi},
	year         = 2024,
	booktitle    = {The Twelfth International Conference on Learning Representations, {ICLR} 2024, Vienna, Austria, May 7-11, 2024},
	publisher    = {OpenReview.net},
	url          = {https://openreview.net/forum?id=wxJ0eXwwda},
	bibsource    = {dblp computer science bibliography, https://dblp.org}
}

@inproceedings{yun2025price,
	title        = {The Price of Format: Diversity Collapse in {LLM}s},
	author       = {Yun, Longfei  and An, Chenyang  and Wang, Zilong  and Peng, Letian  and Shang, Jingbo},
	year         = 2025,
	booktitle    = {Findings of the Association for Computational Linguistics: EMNLP 2025},
	publisher    = {Association for Computational Linguistics},
	address      = {Suzhou, China},
	pages        = {15454--15468},
	doi          = {10.18653/v1/2025.findings-emnlp.836},
	isbn         = {979-8-89176-335-7},
	url          = {https://aclanthology.org/2025.findings-emnlp.836/},
	editor       = {Christodoulopoulos, Christos  and Chakraborty, Tanmoy  and Rose, Carolyn  and Peng, Violet}
}

@misc{zhu2025bare,
	title        = {{BARE}: Leveraging Base Language Models for Few-Shot Synthetic Data Generation},
	author       = {Zhu, Alan and Asawa, Parth and Davis, Jared Quincy and Chen, Lingjiao and Hanin, Boris and Stoica, Ion and Gonzalez, Joseph E. and Zaharia, Matei},
	year         = 2025,
	url          = {https://arxiv.org/abs/2502.01697},
	archiveprefix = {arXiv},
	eprint       = {2502.01697},
	primaryclass = {cs.CL}
}

@inproceedings{west2025base,
	title        = {Base Models Beat Aligned Models at Randomness and Creativity},
	author       = {Peter West and Christopher Potts},
	year         = 2025,
	booktitle    = {Second Conference on Language Modeling},
	url          = {https://openreview.net/forum?id=vqN8uom4A1}
}

@misc{jones2025turing,
	title        = {Large Language Models Pass the {Turing} Test},
	author       = {Jones, Cameron R. and Bergen, Benjamin K.},
	year         = 2025,
	url          = {https://arxiv.org/abs/2503.23674},
	archiveprefix = {arXiv},
	eprint       = {2503.23674},
	primaryclass = {cs.CL}
}

@misc{gandhi2026simulating,
	title        = {Learning to Simulate Human Dialogue},
	author       = {Gandhi, Kanishk and Bhatia, Agam and Goodman, Noah D.},
	year         = 2026,
	url          = {https://arxiv.org/abs/2601.04436},
	archiveprefix = {arXiv},
	eprint       = {2601.04436},
	primaryclass = {cs.CL}
}

@misc{zhu2026dial,
	title        = {DIAL: Direct iterative adversarial learning for realistic multi-turn dialogue simulation},
	author       = {Zhu, Ziyi and Tieleman, Olivier and Stamatis, Caitlin A and Smyth, Luka and Hull, Thomas D and Cahn, Daniel R and Chen, Jinghong and Malgaroli, Matteo},
	year         = 2025,
	url          = {https://arxiv.org/abs/2512.20773},
	archiveprefix = {arXiv},
	eprint       = {2512.20773},
	primaryclass = {cs.CL}
}

@misc{lu2026assistant,
	title        = {The Assistant Axis: Situating and Stabilizing the Default Persona of Language Models},
	author       = {Lu, Christina and Gallagher, Jack and Michala, Jonathan and Fish, Kyle and Lindsey, Jack},
	year         = 2026,
	url          = {https://arxiv.org/abs/2601.10387},
	archiveprefix = {arXiv},
	eprint       = {2601.10387},
	primaryclass = {cs.CL}
}

@article{jelinek1977perplexity,
	title        = {Perplexity---a Measure of the Difficulty of Speech Recognition Tasks},
	author       = {Jelinek, Frederick and Mercer, Robert L. and Bahl, Lalit R. and Baker, James K.},
	year         = 1977,
	journal      = {The Journal of the Acoustical Society of America},
	volume       = 62,
	number       = {S1},
	pages        = {S63--S63},
	doi          = {10.1121/1.2016299}
}

@article{lin1991divergence,
	title        = {Divergence Measures Based on the Shannon Entropy},
	author       = {Lin, Jianhua},
	year         = 1991,
	journal      = {IEEE Transactions on Information Theory},
	volume       = 37,
	number       = 1,
	pages        = {145--151},
	doi          = {10.1109/18.61115}
}

@article{stolcke2000dialogue,
	title        = {Dialogue act modeling for automatic tagging and recognition of conversational speech},
	author       = {Stolcke, Andreas  and Ries, Klaus  and Coccaro, Noah  and Shriberg, Elizabeth  and Bates, Rebecca  and Jurafsky, Daniel  and Taylor, Paul  and Martin, Rachel  and Van Ess-Dykema, Carol  and Meteer, Marie},
	year         = 2000,
	journal      = {Computational Linguistics},
	publisher    = {MIT Press},
	address      = {Cambridge, MA},
	volume       = 26,
	number       = 3,
	pages        = {339--374},
	url          = {https://aclanthology.org/J00-3003/}
}

@inproceedings{papineni2002bleu,
	title        = {{B}leu: a Method for Automatic Evaluation of Machine Translation},
	author       = {Papineni, Kishore  and Roukos, Salim  and Ward, Todd  and Zhu, Wei-Jing},
	year         = 2002,
	booktitle    = {Proceedings of the 40th Annual Meeting of the Association for Computational Linguistics},
	publisher    = {Association for Computational Linguistics},
	address      = {Philadelphia, Pennsylvania, USA},
	pages        = {311--318},
	doi          = {10.3115/1073083.1073135},
	url          = {https://aclanthology.org/P02-1040/},
	editor       = {Isabelle, Pierre  and Charniak, Eugene  and Lin, Dekang}
}

@inproceedings{lin2004rouge,
	title        = {{ROUGE}: A Package for Automatic Evaluation of Summaries},
	author       = {Lin, Chin-Yew},
	year         = 2004,
	booktitle    = {Text Summarization Branches Out},
	publisher    = {Association for Computational Linguistics},
	address      = {Barcelona, Spain},
	pages        = {74--81},
	url          = {https://aclanthology.org/W04-1013/}
}

@inproceedings{li2016diversitypromoting,
	title        = {A Diversity-Promoting Objective Function for Neural Conversation Models},
	author       = {Li, Jiwei  and Galley, Michel  and Brockett, Chris  and Gao, Jianfeng  and Dolan, Bill},
	year         = 2016,
	booktitle    = {Proceedings of the 2016 Conference of the North {A}merican Chapter of the Association for Computational Linguistics: Human Language Technologies},
	publisher    = {Association for Computational Linguistics},
	address      = {San Diego, California},
	pages        = {110--119},
	doi          = {10.18653/v1/N16-1014},
	url          = {https://aclanthology.org/N16-1014/},
	editor       = {Knight, Kevin  and Nenkova, Ani  and Rambow, Owen}
}

@inproceedings{zhu2018texygen,
	title        = {Texygen: {A} Benchmarking Platform for Text Generation Models},
	author       = {Yaoming Zhu and Sidi Lu and Lei Zheng and Jiaxian Guo and Weinan Zhang and Jun Wang and Yong Yu},
	year         = 2018,
	booktitle    = {The 41st International {ACM} {SIGIR} Conference on Research {\&} Development in Information Retrieval, {SIGIR} 2018, Ann Arbor, MI, USA, July 08-12, 2018},
	publisher    = {{ACM}},
	pages        = {1097--1100},
	doi          = {10.1145/3209978.3210080},
	url          = {https://doi.org/10.1145/3209978.3210080},
	bibsource    = {dblp computer science bibliography, https://dblp.org},
	editor       = {Kevyn Collins{-}Thompson and Qiaozhu Mei and Brian D. Davison and Yiqun Liu and Emine Yilmaz}
}

@inproceedings{zhang2020bertscore,
	title        = {BERTScore: Evaluating Text Generation with {BERT}},
	author       = {Tianyi Zhang and Varsha Kishore and Felix Wu and Kilian Q. Weinberger and Yoav Artzi},
	year         = 2020,
	booktitle    = {8th International Conference on Learning Representations, {ICLR} 2020, Addis Ababa, Ethiopia, April 26-30, 2020},
	publisher    = {OpenReview.net},
	url          = {https://openreview.net/forum?id=SkeHuCVFDr},
	bibsource    = {dblp computer science bibliography, https://dblp.org}
}

@article{friedman2023vendi,
	title        = {The Vendi Score: {A} Diversity Evaluation Metric for Machine Learning},
	author       = {Dan Friedman and Adji Bousso Dieng},
	year         = 2023,
	journal      = {Trans. Mach. Learn. Res.},
	volume       = 2023,
	url          = {https://openreview.net/forum?id=g97OHbQyk1},
	bibsource    = {dblp computer science bibliography, https://dblp.org}
}

@inproceedings{li2016persona,
	title        = {A Persona-Based Neural Conversation Model},
	author       = {Li, Jiwei  and Galley, Michel  and Brockett, Chris  and Spithourakis, Georgios  and Gao, Jianfeng  and Dolan, Bill},
	year         = 2016,
	booktitle    = {Proceedings of the 54th Annual Meeting of the Association for Computational Linguistics (Volume 1: Long Papers)},
	publisher    = {Association for Computational Linguistics},
	address      = {Berlin, Germany},
	pages        = {994--1003},
	doi          = {10.18653/v1/P16-1094},
	url          = {https://aclanthology.org/P16-1094/},
	editor       = {Erk, Katrin  and Smith, Noah A.}
}

@inproceedings{zhang2018personalizing,
	title        = {Personalizing Dialogue Agents: {I} have a dog, do you have pets too?},
	author       = {Zhang, Saizheng  and Dinan, Emily  and Urbanek, Jack  and Szlam, Arthur  and Kiela, Douwe  and Weston, Jason},
	year         = 2018,
	booktitle    = {Proceedings of the 56th Annual Meeting of the Association for Computational Linguistics (Volume 1: Long Papers)},
	publisher    = {Association for Computational Linguistics},
	address      = {Melbourne, Australia},
	pages        = {2204--2213},
	doi          = {10.18653/v1/P18-1205},
	url          = {https://aclanthology.org/P18-1205/},
	editor       = {Gurevych, Iryna  and Miyao, Yusuke}
}

@inproceedings{mazare2018training,
	title        = {Training Millions of Personalized Dialogue Agents},
	author       = {Mazar{\'e}, Pierre-Emmanuel  and Humeau, Samuel  and Raison, Martin  and Bordes, Antoine},
	year         = 2018,
	booktitle    = {Proceedings of the 2018 Conference on Empirical Methods in Natural Language Processing},
	publisher    = {Association for Computational Linguistics},
	address      = {Brussels, Belgium},
	pages        = {2775--2779},
	doi          = {10.18653/v1/D18-1298},
	url          = {https://aclanthology.org/D18-1298/},
	editor       = {Riloff, Ellen  and Chiang, David  and Hockenmaier, Julia  and Tsujii, Jun{'}ichi}
}

@article{shanahan2023roleplay,
	title        = {Role Play with Large Language Models},
	author       = {Shanahan, Murray and McDonell, Kyle and Reynolds, Laria},
	year         = 2023,
	journal      = {Nature},
	volume       = 623,
	pages        = {493--498},
	doi          = {10.1038/s41586-023-06647-8}
}

@inproceedings{moon2024virtual,
	title        = {Virtual Personas for Language Models via an Anthology of Backstories},
	author       = {Moon, Suhong  and Abdulhai, Marwa  and Kang, Minwoo  and Suh, Joseph  and Soedarmadji, Widyadewi  and Behar, Eran Kohen  and Chan, David M.},
	year         = 2024,
	booktitle    = {Proceedings of the 2024 Conference on Empirical Methods in Natural Language Processing},
	publisher    = {Association for Computational Linguistics},
	address      = {Miami, Florida, USA},
	pages        = {19864--19897},
	doi          = {10.18653/v1/2024.emnlp-main.1110},
	url          = {https://aclanthology.org/2024.emnlp-main.1110/},
	editor       = {Al-Onaizan, Yaser  and Bansal, Mohit  and Chen, Yun-Nung}
}

@misc{kang2025deep,
	title        = {Deep Binding of Language Model Virtual Personas: A Study on Approximating Political Partisan Misperceptions},
	author       = {Kang, Minwoo and Moon, Suhong and Lee, Seung Hyeong and Raj, Ayush and Suh, Joseph and Chan, David M and Canny, John},
	year         = 2025,
	url          = {https://arxiv.org/abs/2504.11673},
	archiveprefix = {arXiv},
	eprint       = {2504.11673},
	primaryclass = {cs.CL}
}

@misc{moon2026identity,
	title        = {Identity, Cooperation and Framing Effects within Groups of Real and Simulated Humans},
	author       = {Moon, Suhong and Kang, Minwoo and Suh, Joseph and Safdari, Mustafa and Canny, John},
	year         = 2026,
	url          = {https://arxiv.org/abs/2601.16355},
	archiveprefix = {arXiv},
	eprint       = {2601.16355},
	primaryclass = {cs.CL}
}

@inproceedings{wang2024rolellm,
	title        = {{R}ole{LLM}: Benchmarking, Eliciting, and Enhancing Role-Playing Abilities of Large Language Models},
	author       = {Wang, Noah  and Peng, Z.Y.  and Que, Haoran  and Liu, Jiaheng  and Zhou, Wangchunshu  and Wu, Yuhan  and Guo, Hongcheng  and Gan, Ruitong  and Ni, Zehao  and Yang, Jian  and Zhang, Man  and Zhang, Zhaoxiang  and Ouyang, Wanli  and Xu, Ke  and Huang, Wenhao  and Fu, Jie  and Peng, Junran},
	year         = 2024,
	booktitle    = {Findings of the Association for Computational Linguistics: ACL 2024},
	publisher    = {Association for Computational Linguistics},
	address      = {Bangkok, Thailand},
	pages        = {14743--14777},
	doi          = {10.18653/v1/2024.findings-acl.878},
	url          = {https://aclanthology.org/2024.findings-acl.878/},
	editor       = {Ku, Lun-Wei  and Martins, Andre  and Srikumar, Vivek}
}

@inproceedings{park2023generative,
	title        = {Generative Agents: Interactive Simulacra of Human Behavior},
	author       = {Park, Joon Sung and O'Brien, Joseph C. and Cai, Carrie J. and Morris, Meredith Ringel and Liang, Percy and Bernstein, Michael S.},
	year         = 2023,
	booktitle    = {Proceedings of the 36th Annual ACM Symposium on User Interface Software and Technology},
	publisher    = {Association for Computing Machinery},
	series       = {UIST '23},
	doi          = {10.1145/3586183.3606763}
}

@inproceedings{perez2023discovering,
	title        = {Discovering Language Model Behaviors with Model-Written Evaluations},
	author       = {Perez, Ethan  and Ringer, Sam  and Lukosiute, Kamile  and Nguyen, Karina  and Chen, Edwin  and Heiner, Scott  and Pettit, Craig  and Olsson, Catherine  and Kundu, Sandipan  and Kadavath, Saurav  and Jones, Andy  and Chen, Anna  and Mann, Benjamin  and Israel, Brian  and Seethor, Bryan  and McKinnon, Cameron  and Olah, Christopher  and Yan, Da  and Amodei, Daniela  and Amodei, Dario  and Drain, Dawn  and Li, Dustin  and Tran-Johnson, Eli  and Khundadze, Guro  and Kernion, Jackson  and Landis, James  and Kerr, Jamie  and Mueller, Jared  and Hyun, Jeeyoon  and Landau, Joshua  and Ndousse, Kamal  and Goldberg, Landon  and Lovitt, Liane  and Lucas, Martin  and Sellitto, Michael  and Zhang, Miranda  and Kingsland, Neerav  and Elhage, Nelson  and Joseph, Nicholas  and Mercado, Noemi  and DasSarma, Nova  and Rausch, Oliver  and Larson, Robin  and McCandlish, Sam  and Johnston, Scott  and Kravec, Shauna  and El Showk, Sheer  and Lanham, Tamera  and Telleen-Lawton, Timothy  and Brown, Tom  and Henighan, Tom  and Hume, Tristan  and Bai, Yuntao  and Hatfield-Dodds, Zac  and Clark, Jack  and Bowman, Samuel R.  and Askell, Amanda  and Grosse, Roger  and Hernandez, Danny  and Ganguli, Deep  and Hubinger, Evan  and Schiefer, Nicholas  and Kaplan, Jared},
	year         = 2023,
	booktitle    = {Findings of the Association for Computational Linguistics: ACL 2023},
	publisher    = {Association for Computational Linguistics},
	address      = {Toronto, Canada},
	pages        = {13387--13434},
	doi          = {10.18653/v1/2023.findings-acl.847},
	url          = {https://aclanthology.org/2023.findings-acl.847/},
	editor       = {Rogers, Anna  and Boyd-Graber, Jordan  and Okazaki, Naoaki}
}

@inproceedings{sharma2024towards,
	title        = {Towards Understanding Sycophancy in Language Models},
	author       = {Mrinank Sharma and Meg Tong and Tomasz Korbak and David Duvenaud and Amanda Askell and Samuel R. Bowman and Esin Durmus and Zac Hatfield{-}Dodds and Scott R. Johnston and Shauna Kravec and Timothy Maxwell and Sam McCandlish and Kamal Ndousse and Oliver Rausch and Nicholas Schiefer and Da Yan and Miranda Zhang and Ethan Perez},
	year         = 2024,
	booktitle    = {The Twelfth International Conference on Learning Representations, {ICLR} 2024, Vienna, Austria, May 7-11, 2024},
	publisher    = {OpenReview.net},
	url          = {https://openreview.net/forum?id=tvhaxkMKAn},
	bibsource    = {dblp computer science bibliography, https://dblp.org}
}

@inproceedings{kirk2024understanding,
	title        = {Understanding the Effects of {RLHF} on {LLM} Generalisation and Diversity},
	author       = {Robert Kirk and Ishita Mediratta and Christoforos Nalmpantis and Jelena Luketina and Eric Hambro and Edward Grefenstette and Roberta Raileanu},
	year         = 2024,
	booktitle    = {The Twelfth International Conference on Learning Representations, {ICLR} 2024, Vienna, Austria, May 7-11, 2024},
	publisher    = {OpenReview.net},
	url          = {https://openreview.net/forum?id=PXD3FAVHJT},
	bibsource    = {dblp computer science bibliography, https://dblp.org}
}

@inproceedings{padmakumar2024does,
	title        = {Does Writing with Language Models Reduce Content Diversity?},
	author       = {Vishakh Padmakumar and He He},
	year         = 2024,
	booktitle    = {The Twelfth International Conference on Learning Representations, {ICLR} 2024, Vienna, Austria, May 7-11, 2024},
	publisher    = {OpenReview.net},
	url          = {https://openreview.net/forum?id=Feiz5HtCD0},
	bibsource    = {dblp computer science bibliography, https://dblp.org}
}

@inproceedings{li2024measuring,
	title        = {Measuring and Controlling Instruction (In)Stability in Language Model Dialogs},
	author       = {Li, Kenneth and Liu, Tianle and Bashkansky, Naomi and Bau, David and Vi{\'e}gas, Fernanda and Pfister, Hanspeter and Wattenberg, Martin},
	year         = 2024,
	booktitle    = {Proceedings of the First Conference on Language Modeling},
	note         = {arXiv:2402.10962}
}

@article{argyle2023out,
	title        = {Out of One, Many: Using Language Models to Simulate Human Samples},
	author       = {Argyle, Lisa P. and Busby, Ethan C. and Fulda, Nancy and Gubler, Joshua R. and Rytting, Christopher and Wingate, David},
	year         = 2023,
	journal      = {Political Analysis},
	volume       = 31,
	number       = 3,
	pages        = {337--351},
	doi          = {10.1017/pan.2023.2}
}

@inproceedings{tseng2024two,
	title        = {Two Tales of Persona in {LLM}s: A Survey of Role-Playing and Personalization},
	author       = {Tseng, Yu-Min  and Huang, Yu-Chao  and Hsiao, Teng-Yun  and Chen, Wei-Lin  and Huang, Chao-Wei  and Meng, Yu  and Chen, Yun-Nung},
	year         = 2024,
	booktitle    = {Findings of the Association for Computational Linguistics: EMNLP 2024},
	publisher    = {Association for Computational Linguistics},
	address      = {Miami, Florida, USA},
	pages        = {16612--16631},
	doi          = {10.18653/v1/2024.findings-emnlp.969},
	url          = {https://aclanthology.org/2024.findings-emnlp.969/},
	editor       = {Al-Onaizan, Yaser  and Bansal, Mohit  and Chen, Yun-Nung}
}

@article{schatzmann2006survey,
	title        = {A Survey of Statistical User Simulation Techniques for Reinforcement-Learning of Dialogue Management Strategies},
	author       = {Schatzmann, Jost and Weilhammer, Karl and Stuttle, Matthew N. and Young, Steve},
	year         = 2006,
	journal      = {The Knowledge Engineering Review},
	volume       = 21,
	number       = 2,
	pages        = {97--126},
	doi          = {10.1017/S0269888906000944}
}

@inproceedings{schatzmann2007agenda,
	title        = {Agenda-Based User Simulation for Bootstrapping a {POMDP} Dialogue System},
	author       = {Schatzmann, Jost  and Thomson, Blaise  and Weilhammer, Karl  and Ye, Hui  and Young, Steve},
	year         = 2007,
	booktitle    = {Human Language Technologies 2007: The Conference of the North {A}merican Chapter of the Association for Computational Linguistics; Companion Volume, Short Papers},
	publisher    = {Association for Computational Linguistics},
	address      = {Rochester, New York},
	pages        = {149--152},
	url          = {https://aclanthology.org/N07-2038/},
	editor       = {Sidner, Candace  and Schultz, Tanja  and Stone, Matthew  and Zhai, ChengXiang}
}

@inproceedings{kreyssig2018neural,
	title        = {Neural User Simulation for Corpus-based Policy Optimisation of Spoken Dialogue Systems},
	author       = {Kreyssig, Florian  and Casanueva, I{\~n}igo  and Budzianowski, Pawe{\l}  and Ga{\v{s}}i{\'c}, Milica},
	year         = 2018,
	booktitle    = {Proceedings of the 19th Annual {SIG}dial Meeting on Discourse and Dialogue},
	publisher    = {Association for Computational Linguistics},
	address      = {Melbourne, Australia},
	pages        = {60--69},
	doi          = {10.18653/v1/W18-5007},
	url          = {https://aclanthology.org/W18-5007/},
	editor       = {Komatani, Kazunori  and Litman, Diane  and Yu, Kai  and Papangelis, Alex  and Cavedon, Lawrence  and Nakano, Mikio}
}

@inproceedings{lin2022gentus,
	title        = {{G}en{TUS}: Simulating User Behaviour and Language in Task-oriented Dialogues with Generative Transformers},
	author       = {Lin, Hsien-chin  and Geishauser, Christian  and Feng, Shutong  and Lubis, Nurul  and van Niekerk, Carel  and Heck, Michael  and Ga{\v{s}}i{\'c}, Milica},
	year         = 2022,
	booktitle    = {Proceedings of the 23rd Annual Meeting of the Special Interest Group on Discourse and Dialogue},
	publisher    = {Association for Computational Linguistics},
	address      = {Edinburgh, UK},
	pages        = {270--282},
	doi          = {10.18653/v1/2022.sigdial-1.28},
	url          = {https://aclanthology.org/2022.sigdial-1.28/},
	editor       = {Lemon, Oliver  and Hakkani-Tur, Dilek  and Li, Junyi Jessy  and Ashrafzadeh, Arash  and Garcia, Daniel Hern{\'a}ndez  and Alikhani, Malihe  and Vandyke, David  and Du{\v{s}}ek, Ond{\v{r}}ej}
}

@misc{terragni2023incontext,
	title        = {In-Context Learning User Simulators for Task-Oriented Dialog Systems},
	author       = {Terragni, Silvia and Filipavicius, Modestas and Khau, Nghia and Guedes, Bruna and Manso, Andr{\'e} and Mathis, Roland},
	year         = 2023,
	url          = {https://arxiv.org/abs/2306.00774},
	archiveprefix = {arXiv},
	eprint       = {2306.00774},
	primaryclass = {cs.CL}
}

@inproceedings{sekulic2024reliable,
	title        = {Reliable {LLM}-based User Simulator for Task-Oriented Dialogue Systems},
	author       = {Sekulic, Ivan  and Terragni, Silvia  and Guimar{\~a}es, Victor  and Khau, Nghia  and Guedes, Bruna  and Filipavicius, Modestas  and Manso, Andre Ferreira  and Mathis, Roland},
	year         = 2024,
	booktitle    = {Proceedings of the 1st Workshop on Simulating Conversational Intelligence in Chat (SCI-CHAT 2024)},
	publisher    = {Association for Computational Linguistics},
	address      = {St. Julians, Malta},
	pages        = {19--35},
	doi          = {10.18653/v1/2024.scichat-1.3},
	url          = {https://aclanthology.org/2024.scichat-1.3/},
	editor       = {Graham, Yvette  and Liu, Qun  and Lampouras, Gerasimos  and Iacobacci, Ignacio  and Madden, Sinead  and Khalid, Haider  and Qureshi, Rameez}
}

@article{radford2019language,
	title        = {Language Models are Unsupervised Multitask Learners},
	author       = {Radford, Alec and Wu, Jeffrey and Child, Rewon and Luan, David and Amodei, Dario and Sutskever, Ilya},
	year         = 2019,
	journal      = {OpenAI Technical Report},
	url          = {https://cdn.openai.com/better-language-models/language_models_are_unsupervised_multitask_learners.pdf}
}

@inproceedings{brown2020language,
	title        = {Language Models are Few-Shot Learners},
	author       = {Tom B. Brown and Benjamin Mann and Nick Ryder and Melanie Subbiah and Jared Kaplan and Prafulla Dhariwal and Arvind Neelakantan and Pranav Shyam and Girish Sastry and Amanda Askell and Sandhini Agarwal and Ariel Herbert{-}Voss and Gretchen Krueger and Tom Henighan and Rewon Child and Aditya Ramesh and Daniel M. Ziegler and Jeffrey Wu and Clemens Winter and Christopher Hesse and Mark Chen and Eric Sigler and Mateusz Litwin and Scott Gray and Benjamin Chess and Jack Clark and Christopher Berner and Sam McCandlish and Alec Radford and Ilya Sutskever and Dario Amodei},
	year         = 2020,
	booktitle    = {Advances in Neural Information Processing Systems 33: Annual Conference on Neural Information Processing Systems 2020, NeurIPS 2020, December 6-12, 2020, virtual},
	url          = {https://proceedings.neurips.cc/paper/2020/hash/1457c0d6bfcb4967418bfb8ac142f64a-Abstract.html},
	bibsource    = {dblp computer science bibliography, https://dblp.org},
	editor       = {Hugo Larochelle and Marc'Aurelio Ranzato and Raia Hadsell and Maria{-}Florina Balcan and Hsuan{-}Tien Lin}
}

@inproceedings{lewis2017deal,
	title        = {Deal or No Deal? End-to-End Learning of Negotiation Dialogues},
	author       = {Lewis, Mike  and Yarats, Denis  and Dauphin, Yann  and Parikh, Devi  and Batra, Dhruv},
	year         = 2017,
	booktitle    = {Proceedings of the 2017 Conference on Empirical Methods in Natural Language Processing},
	publisher    = {Association for Computational Linguistics},
	address      = {Copenhagen, Denmark},
	pages        = {2443--2453},
	doi          = {10.18653/v1/D17-1259},
	url          = {https://aclanthology.org/D17-1259/},
	editor       = {Palmer, Martha  and Hwa, Rebecca  and Riedel, Sebastian}
}

@inproceedings{li2023camel,
	title        = {{CAMEL:} Communicative Agents for "Mind" Exploration of Large Language Model Society},
	author       = {Guohao Li and Hasan Hammoud and Hani Itani and Dmitrii Khizbullin and Bernard Ghanem},
	year         = 2023,
	booktitle    = {Advances in Neural Information Processing Systems 36: Annual Conference on Neural Information Processing Systems 2023, NeurIPS 2023, New Orleans, LA, USA, December 10 - 16, 2023},
	url          = {http://papers.nips.cc/paper\_files/paper/2023/hash/a3621ee907def47c1b952ade25c67698-Abstract-Conference.html},
	bibsource    = {dblp computer science bibliography, https://dblp.org},
	editor       = {Alice Oh and Tristan Naumann and Amir Globerson and Kate Saenko and Moritz Hardt and Sergey Levine}
}

@inproceedings{andreas2022language,
	title        = {Language Models as Agent Models},
	author       = {Andreas, Jacob},
	year         = 2022,
	booktitle    = {Findings of the Association for Computational Linguistics: EMNLP 2022},
	publisher    = {Association for Computational Linguistics},
	address      = {Abu Dhabi, United Arab Emirates},
	pages        = {5769--5779},
	doi          = {10.18653/v1/2022.findings-emnlp.423},
	url          = {https://aclanthology.org/2022.findings-emnlp.423/},
	editor       = {Goldberg, Yoav  and Kozareva, Zornitsa  and Zhang, Yue}
}

@inproceedings{santurkar2023whose,
	title        = {Whose Opinions Do Language Models Reflect?},
	author       = {Shibani Santurkar and Esin Durmus and Faisal Ladhak and Cinoo Lee and Percy Liang and Tatsunori Hashimoto},
	year         = 2023,
	booktitle    = {International Conference on Machine Learning, {ICML} 2023, 23-29 July 2023, Honolulu, Hawaii, {USA}},
	publisher    = {{PMLR}},
	series       = {Proceedings of Machine Learning Research},
	volume       = 202,
	pages        = {29971--30004},
	url          = {https://proceedings.mlr.press/v202/santurkar23a.html},
	bibsource    = {dblp computer science bibliography, https://dblp.org},
	editor       = {Andreas Krause and Emma Brunskill and Kyunghyun Cho and Barbara Engelhardt and Sivan Sabato and Jonathan Scarlett}
}

@misc{park2024generativeagentsimulations1000,
	title        = {LLM agents grounded in self-reports enable general-purpose simulation of individuals},
	author       = {Park, Joon Sung and Zou, Carolyn Q and Kamphorst, Jonne and Egan, Niles and Shaw, Aaron and Hill, Benjamin Mako and Cai, Carrie and Morris, Meredith Ringel and Liang, Percy and Willer, Robb and others},
	year         = 2024,
	url          = {https://arxiv.org/abs/2411.10109},
	archiveprefix = {arXiv},
	eprint       = {2411.10109},
	primaryclass = {cs.AI}
}

@inproceedings{aher2023using,
	title        = {Using Large Language Models to Simulate Multiple Humans and Replicate Human Subject Studies},
	author       = {Gati V. Aher and Rosa I. Arriaga and Adam Tauman Kalai},
	year         = 2023,
	booktitle    = {International Conference on Machine Learning, {ICML} 2023, 23-29 July 2023, Honolulu, Hawaii, {USA}},
	publisher    = {{PMLR}},
	series       = {Proceedings of Machine Learning Research},
	volume       = 202,
	pages        = {337--371},
	url          = {https://proceedings.mlr.press/v202/aher23a.html},
	bibsource    = {dblp computer science bibliography, https://dblp.org},
	editor       = {Andreas Krause and Emma Brunskill and Kyunghyun Cho and Barbara Engelhardt and Sivan Sabato and Jonathan Scarlett}
}

@inproceedings{simmons2022moral,
	title        = {Moral Mimicry: Large Language Models Produce Moral Rationalizations Tailored to Political Identity},
	author       = {Simmons, Gabriel},
	year         = 2023,
	booktitle    = {Proceedings of the 61st Annual Meeting of the Association for Computational Linguistics (Volume 4: Student Research Workshop)},
	publisher    = {Association for Computational Linguistics},
	address      = {Toronto, Canada},
	pages        = {282--297},
	doi          = {10.18653/v1/2023.acl-srw.40},
	url          = {https://aclanthology.org/2023.acl-srw.40/},
	editor       = {Padmakumar, Vishakh  and Vallejo, Gisela  and Fu, Yao}
}

@article{ziems2023large,
	title        = {Can Large Language Models Transform Computational Social Science?},
	author       = {Ziems, Caleb  and Held, William  and Shaikh, Omar  and Chen, Jiaao  and Zhang, Zhehao  and Yang, Diyi},
	year         = 2024,
	journal      = {Computational Linguistics},
	publisher    = {MIT Press},
	address      = {Cambridge, MA},
	volume       = 50,
	number       = 1,
	pages        = {237--291},
	doi          = {10.1162/coli_a_00502},
	url          = {https://aclanthology.org/2024.cl-1.8/}
}

@article{bail2024generative,
	title        = {Can Generative {AI} Improve Social Science?},
	author       = {Bail, Christopher A.},
	year         = 2024,
	journal      = {Proceedings of the National Academy of Sciences},
	volume       = 121,
	number       = 21,
	pages        = {e2314021121},
	doi          = {10.1073/pnas.2314021121}
}

@inproceedings{jiang2025hivemind,
	title        = {Artificial Hivemind: The Open-Ended Homogeneity of Language Models (and Beyond)},
	author       = {Liwei Jiang and Yuanjun Chai and Margaret Li and Mickel Liu and Raymond Fok and Nouha Dziri and Yulia Tsvetkov and Maarten Sap and Yejin Choi},
	year         = 2025,
	booktitle    = {Advances in Neural Information Processing Systems 38: Annual Conference on Neural Information Processing Systems 2025, NeurIPS 2025, San Diego, CA, USA, December 2-7, 2025 / Mexico City, Mexico, November 30 - December 5, 2025},
	url          = {http://papers.nips.cc/paper\_files/paper/2025/hash/754d5a526a5ee5a47220664a0eb92751-Abstract-Datasets\_and\_Benchmarks\_Track.html},
	bibsource    = {dblp computer science bibliography, https://dblp.org},
	editor       = {Danielle Belgrave and Cheng Zhang and Laura N. Montoya and Hsuan{-}Tien Lin and Razvan Pascanu and Piotr Koniusz and Marzyeh Ghassemi and Nancy Chen and Iv{\'{a}}n Vladimir Meza Ru{\'{\i}}z and Arturo Loaiza{-}Bonilla}
}

@misc{zhou2026sim2real,
	title        = {Mind the Sim2Real Gap in User Simulation for Agentic Tasks},
	author       = {Zhou, Xuhui and Sun, Weiwei and Ma, Qianou and Xie, Yiqing and Liu, Jiarui and Du, Weihua and Welleck, Sean and Yang, Yiming and Neubig, Graham and Wu, Sherry Tongshuang and Sap, Maarten},
	year         = 2026,
	url          = {https://arxiv.org/abs/2603.11245},
	archiveprefix = {arXiv},
	eprint       = {2603.11245},
	primaryclass = {cs.AI}
}

@inproceedings{maitra2025dialogue,
	title        = {Dialogue Acts as a Lens on Human{--}{LLM} Interaction: Analyzing Conversational Norms in Model-Generated Responses},
	author       = {Maitra, Arunima  and French, Dorothea  and von der Wense, Katharina},
	year         = 2025,
	booktitle    = {Proceedings of the Fourth Workshop on Bridging Human-Computer Interaction and Natural Language Processing (HCI+NLP)},
	publisher    = {Association for Computational Linguistics},
	address      = {Suzhou, China},
	pages        = {317--325},
	doi          = {10.18653/v1/2025.hcinlp-1.25},
	isbn         = {979-8-89176-353-1},
	url          = {https://aclanthology.org/2025.hcinlp-1.25/},
	editor       = {Blodgett, Su Lin  and Curry, Amanda Cercas  and Dev, Sunipa  and Li, Siyan  and Madaio, Michael  and Wang, Jack  and Wu, Sherry Tongshuang  and Xiao, Ziang  and Yang, Diyi}
}

@inproceedings{shaikh2024grounding,
	title        = {Grounding Gaps in Language Model Generations},
	author       = {Shaikh, Omar  and Gligori{\'c}, Kristina  and Khetan, Ashna  and Gerstgrasser, Matthias  and Yang, Diyi  and Jurafsky, Dan},
	year         = 2024,
	booktitle    = {Proceedings of the 2024 Conference of the North American Chapter of the Association for Computational Linguistics: Human Language Technologies (Volume 1: Long Papers)},
	publisher    = {Association for Computational Linguistics},
	address      = {Mexico City, Mexico},
	pages        = {6279--6296},
	doi          = {10.18653/v1/2024.naacl-long.348},
	url          = {https://aclanthology.org/2024.naacl-long.348/},
	editor       = {Duh, Kevin  and Gomez, Helena  and Bethard, Steven}
}

@misc{emi2024technicalreportpangramaigenerated,
	title        = {Technical Report on the Pangram AI-Generated Text Classifier},
	author       = {Bradley Emi and Max Spero},
	year         = 2024,
	url          = {https://arxiv.org/abs/2402.14873},
	archiveprefix = {arXiv},
	eprint       = {2402.14873},
	primaryclass = {cs.CL}
}

@misc{zhang2026userlmr1modelinghumanreasoning,
	title        = {UserLM-R1: Modeling Human Reasoning in User Language Models with Multi-Reward Reinforcement Learning},
	author       = {Feng Zhang and Shijia Li and Chunmao Zhang and Zhanyu Ma and Jun Xu and Jiuchong Gao and Jinghua Hao and Renqing He and Jingwen Xu and Han Liu},
	year         = 2026,
	url          = {https://arxiv.org/abs/2601.09215},
	archiveprefix = {arXiv},
	eprint       = {2601.09215},
	primaryclass = {cs.CL}
}

@article{cheng2026sycophantic,
	title        = {Sycophantic AI decreases prosocial intentions and promotes dependence},
	author       = {Cheng, Myra and Lee, Cinoo and Khadpe, Pranav and Yu, Sunny and Han, Dyllan and Jurafsky, Dan},
	year         = 2026,
	journal      = {Science},
	publisher    = {American Association for the Advancement of Science},
	volume       = 391,
	number       = 6792,
	pages        = {eaec8352}
}

@inproceedings{anthis2025social,
	title        = {Position: LLM social simulations are a promising research method},
	author       = {Anthis, Jacy Reese and Liu, Ryan and Richardson, Sean M and Kozlowski, Austin C and Koch, Bernard and Brynjolfsson, Erik and Evans, James and Bernstein, Michael S},
	year         = 2025,
	booktitle    = {Forty-second International Conference on Machine Learning Position Paper Track}
}

@inproceedings{li2025llm,
	title        = {{LLM} Generated Persona is a Promise with a Catch},
	author       = {Leon Li and Haozhe Chen and Hongseok Namkoong and Tianyi Peng},
	year         = 2025,
	booktitle    = {Advances in Neural Information Processing Systems 38: Annual Conference on Neural Information Processing Systems 2025, NeurIPS 2025, San Diego, CA, USA, December 2-7, 2025 / Mexico City, Mexico, November 30 - December 5, 2025},
	url          = {http://papers.nips.cc/paper\_files/paper/2025/hash/5fc150b75e8b509735564de16bc3527b-Abstract-Position\_Paper\_Track.html},
	bibsource    = {dblp computer science bibliography, https://dblp.org},
	editor       = {Danielle Belgrave and Cheng Zhang and Laura N. Montoya and Hsuan{-}Tien Lin and Razvan Pascanu and Piotr Koniusz and Marzyeh Ghassemi and Nancy Chen and Iv{\'{a}}n Vladimir Meza Ru{\'{\i}}z and Arturo Loaiza{-}Bonilla}
}

@article{lyman2025balancing,
	title        = {Balancing large language model alignment and algorithmic fidelity in social science research},
	author       = {Alex Lyman and Bryce Hepner and Lisa P Argyle and Ethan C Busby and Joshua R Gubler and David Wingate},
	year         = 2025,
	journal      = {Sociological Methods and Research},
	publisher    = {SAGE Publications},
	volume       = 54,
	issue        = 3
}

@misc{yao2024tau,
	title        = {$tau$-bench: A Benchmark for Tool-Agent-User Interaction in Real-World Domains},
	author       = {Yao, Shunyu and Shinn, Noah and Razavi, Pedram and Narasimhan, Karthik},
	year         = 2024,
	url          = {https://arxiv.org/abs/2406.12045},
	archiveprefix = {arXiv},
	eprint       = {2406.12045},
	primaryclass = {cs.AI}
}

@inproceedings{ni2026survey,
	title        = {A Survey on {LLM}-based Conversational User Simulation},
	author       = {Ni, Bo  and Wang, Yu  and Wang, Leyao  and Kveton, Branislav  and Dernoncourt, Franck  and Xia, Yu  and Chen, Hongjie  and Luera, Reuben  and Basu, Samyadeep  and Mukherjee, Subhojyoti  and Mathur, Puneet  and Ahmed, Nesreen K.  and Wu, Junda  and Li, Li  and Zhang, Huixin  and Zhang, Ruiyi  and Yu, Tong  and Kim, Sungchul  and Gu, Jiuxiang  and Tu, Zhengzhong  and Siu, Alexa  and Wang, Zichao  and Yoon, Seunghyun  and Lipka, Nedim  and Park, Namyong  and Lin, Zihao  and Bui, Trung  and Zhao, Yue  and Derr, Tyler  and Rossi, Ryan A.},
	year         = 2026,
	booktitle    = {Proceedings of the 19th Conference of the {E}uropean Chapter of the {A}ssociation for {C}omputational {L}inguistics (Volume 1: Long Papers)},
	publisher    = {Association for Computational Linguistics},
	address      = {Rabat, Morocco},
	pages        = {4266--4301},
	doi          = {10.18653/v1/2026.eacl-long.200},
	isbn         = {979-8-89176-380-7},
	url          = {https://aclanthology.org/2026.eacl-long.200/},
	editor       = {Demberg, Vera  and Inui, Kentaro  and Marquez, Llu{\'i}s}
}

@inproceedings{ivey2024real,
	title        = {Real or Robotic? Assessing Whether {LLM}s Accurately Simulate Qualities of Human Responses in Human-{LLM} Dialogue},
	author       = {Ivey, Jonathan  and Kumar, Shivani  and Liu, Jiayu  and Shen, Hua  and Rakshit, Sushrita  and Raju, Rohan  and Zhang, Haotian  and Ananthasubramaniam, Aparna  and Kim, Junghwan  and Yi, Bowen  and Wright, Dustin  and Israeli, Abraham  and M{\o}ller, Anders Giovanni  and Zhang, Lechen  and Jurgens, David},
	year         = 2026,
	booktitle    = {Findings of the {A}ssociation for {C}omputational {L}inguistics: {ACL} 2026},
	publisher    = {Association for Computational Linguistics},
	address      = {San Diego, California, United States},
	pages        = {41395--41432},
	doi          = {10.18653/v1/2026.findings-acl.2060},
	isbn         = {979-8-89176-395-1},
	url          = {https://aclanthology.org/2026.findings-acl.2060/},
	editor       = {Liakata, Maria  and Moreira, Viviane P.  and Zhang, Jiajun  and Jurgens, David}
}

@inproceedings{georgila2006user,
	title        = {User Simulation for Spoken Dialogue Systems: Learning and Evaluation},
	author       = {Georgila, Kallirroi and Henderson, James and Lemon, Oliver},
	year         = 2006,
	booktitle    = {Proceedings of the 9th International Conference on Spoken Language Processing (INTERSPEECH-ICSLP)},
	address      = {Pittsburgh, USA},
	pages        = {1065--1068},
	doi          = {10.21437/Interspeech.2006-160}
}

@article{crockett2025ai,
	title        = {AI Surrogates and illusions of generalizability in cognitive science},
	author       = {Crockett, Molly J and Messeri, Lisa},
	year         = 2025,
	journal      = {Trends in Cognitive Sciences},
	publisher    = {Elsevier}
}

@inproceedings{kapania2025simulacrum,
	title        = {Simulacrum of Stories: Examining Large Language Models as Qualitative Research Participants},
	author       = {Shivani Kapania and William Agnew and Motahhare Eslami and Hoda Heidari and Sarah E. Fox},
	year         = 2025,
	booktitle    = {Proceedings of the 2025 {CHI} Conference on Human Factors in Computing Systems, {CHI} 2025, YokohamaJapan, 26 April 2025- 1 May 2025},
	publisher    = {{ACM}},
	pages        = {489:1--489:17},
	doi          = {10.1145/3706598.3713220},
	url          = {https://doi.org/10.1145/3706598.3713220},
	bibsource    = {dblp computer science bibliography, https://dblp.org},
	editor       = {Naomi Yamashita and Vanessa Evers and Koji Yatani and Sharon Xianghua Ding and Bongshin Lee and Marshini Chetty and Phoebe O. Toups Dugas}
}

@inproceedings{baidya2025behavior,
	title        = {The Behavior Gap: Evaluating Zero-shot {LLM} Agents in Complex Task-Oriented Dialogs},
	author       = {Baidya, Avinash  and Das, Kamalika  and Gao, Xiang},
	year         = 2025,
	booktitle    = {Findings of the Association for Computational Linguistics: ACL 2025},
	publisher    = {Association for Computational Linguistics},
	address      = {Vienna, Austria},
	pages        = {23455--23472},
	doi          = {10.18653/v1/2025.findings-acl.1205},
	isbn         = {979-8-89176-256-5},
	url          = {https://aclanthology.org/2025.findings-acl.1205/},
	editor       = {Che, Wanxiang  and Nabende, Joyce  and Shutova, Ekaterina  and Pilehvar, Mohammad Taher}
}

@inproceedings{shaikh2025navigating,
	title        = {Navigating Rifts in Human-{LLM} Grounding: Study and Benchmark},
	author       = {Shaikh, Omar  and Mozannar, Hussein  and Bansal, Gagan  and Fourney, Adam  and Horvitz, Eric},
	year         = 2025,
	booktitle    = {Proceedings of the 63rd Annual Meeting of the Association for Computational Linguistics (Volume 1: Long Papers)},
	publisher    = {Association for Computational Linguistics},
	address      = {Vienna, Austria},
	pages        = {20832--20847},
	doi          = {10.18653/v1/2025.acl-long.1016},
	isbn         = {979-8-89176-251-0},
	url          = {https://aclanthology.org/2025.acl-long.1016/},
	editor       = {Che, Wanxiang  and Nabende, Joyce  and Shutova, Ekaterina  and Pilehvar, Mohammad Taher}
}

@inproceedings{yu2026pragmatic,
	title        = {The Pragmatic Mind of Machines: Tracing the Emergence of Pragmatic Competence in Large Language Models},
	author       = {Yu, Kefan  and Zeng, Qingcheng  and Xuan, Weihao  and Li, Wanxin  and Wu, Jingyi  and Voigt, Rob},
	year         = 2026,
	booktitle    = {Proceedings of the 19th Conference of the {E}uropean Chapter of the {A}ssociation for {C}omputational {L}inguistics (Volume 1: Long Papers)},
	publisher    = {Association for Computational Linguistics},
	address      = {Rabat, Morocco},
	pages        = {192--213},
	doi          = {10.18653/v1/2026.eacl-long.9},
	isbn         = {979-8-89176-380-7},
	url          = {https://aclanthology.org/2026.eacl-long.9/},
	editor       = {Demberg, Vera  and Inui, Kentaro  and Marquez, Llu{\'i}s}
}

@article{levin2000stochastic,
	title        = {A stochastic model of human-machine interaction for learning dialog strategies},
	author       = {Levin, Esther and Pieraccini, Roberto and Eckert, Wieland and others},
	year         = 2000,
	journal      = {IEEE Transactions on speech and audio processing},
	volume       = 8,
	number       = 1,
	pages        = {11--23}
}

@article{bender2021dangers,
	title        = {On the dangers of stochastic parrots: Can language models be too big?},
	author       = {Emily M Bender and Timnit Gebru and Angelina McMillan-Major and Shmargaret Shmitchell},
	year         = 2021,
	journal      = {ACM conference on fairness, accountability, and transparency}
}

@misc{yu2024finetuninglanguagemodelsgenerative,
	title        = {Fine-tuning Language Models with Generative Adversarial Reward Modelling},
	author       = {Zhang Ze Yu and Lau Jia Jaw and Zhang Hui and Bryan Kian Hsiang Low},
	year         = 2023,
	url          = {https://arxiv.org/abs/2305.06176},
	archiveprefix = {arXiv},
	eprint       = {2305.06176},
	primaryclass = {cs.CL}
}

@inproceedings{wu2026largelanguagemodelssensitive,
	title        = {Are Large Language Models Sensitive to the Motives Behind Communication?},
	author       = {Addison J. Wu and Ryan Liu and Kerem Oktar and Theodore R. Sumers and Tom Griffiths},
	year         = 2025,
	booktitle    = {Advances in Neural Information Processing Systems 38: Annual Conference on Neural Information Processing Systems 2025, NeurIPS 2025, San Diego, CA, USA, December 2-7, 2025 / Mexico City, Mexico, November 30 - December 5, 2025},
	url          = {http://papers.nips.cc/paper\_files/paper/2025/hash/e5fb1d7e7c1541eb0b6132b7839baf34-Abstract-Conference.html},
	bibsource    = {dblp computer science bibliography, https://dblp.org},
	editor       = {Danielle Belgrave and Cheng Zhang and Laura N. Montoya and Hsuan{-}Tien Lin and Razvan Pascanu and Piotr Koniusz and Marzyeh Ghassemi and Nancy Chen and Iv{\'{a}}n Vladimir Meza Ru{\'{\i}}z and Arturo Loaiza{-}Bonilla}
}

@misc{binz2026posttrainingmakeslargelanguage,
	title        = {Post-training makes large language models less human-like},
	author       = {Marcel Binz and Elif Akata and Abdullah Almaatouq and Mohammed Alsobay and Oleksii Ariasov and Franziska Brändle and David Broska and Jason W. Burton and Nuno Busch and Frederick Callaway and Vanessa Cheung and Brian Christian and Julian Coda-Forno and Can Demircan and Vittoria Dentella and Maria K. Eckstein and Noémi Éltető and Michael Franke and Thomas L. Griffiths and Fritz Günther and Susanne Haridi and Sebastian Hellmann and Stefan Herytash and Linus Hof and Eleanor Holton and Isabelle Hoxha and Zak Hussain and Akshay Jagadish and Elif Kara and Valentin Kriegmair and Evelina Leivada and Li Ji-An and Tobias Ludwig and Maximilian Maier and Marcelo G. Mattar and Marvin Mathony and Alireza Modirshanechi and Robin Na and Mariia Nadverniuk and Antonios Nasioulas and Surabhi S. Nath and Helen Niemeyer and Kate Nussenbaum and Sebastian Olschewski and Thorsten Pachur and Stefano Palminteri and Aliona Petrenco and Camille V. Phaneuf-Hadd and Angelo Pirrone and Manuel Rausch and Laura Raveling and Shashank Reddy and Milena Rmus and Evan M. Russek and Tankred Saanum and Kai Sandbrink and Louis Schiekiera and Johannes A. Schubert and Luca M. Schulze Buschoff and Nishad Singhi and Leah H. Somerville and Mikhail S. Spektor and Xin Sui and Christopher Summerfield and Mirko Thalmann and Anna I. Thoma and Taisiia Tikhomirova and Vuong Truong and Polina Tsvilodub and Konstantinos Voudouris and Kristin Witte and Shuchen Wu and Dirk U. Wulff and Hua-Dong Xiong and Songlin Xu and Lance Ying and Xinyu Zhang and Jian-Qiao Zhu and Eric Schulz},
	year         = 2026,
	url          = {https://arxiv.org/abs/2605.07632},
	archiveprefix = {arXiv},
	eprint       = {2605.07632},
	primaryclass = {cs.CL}
}

@misc{lin2026illusioninterventionllmsimulatedexperiment,
	title        = {The Illusion of Intervention: Your LLM-Simulated Experiment is an Observational Study},
	author       = {Victoria Lin and Taedong Yun and Maja Matarić and John Canny and Arthur Gretton and Alexander D'Amour},
	year         = 2026,
	url          = {https://arxiv.org/abs/2605.20767},
	archiveprefix = {arXiv},
	eprint       = {2605.20767},
	primaryclass = {cs.CL}
}
